\documentclass{article} 
\usepackage{iclr2024_conference,times}

\usepackage[resetlabels]{multibib}
\newcites{app}{References}

\usepackage[utf8]{inputenc} 
\usepackage[T1]{fontenc}    
\usepackage{hyperref}
\usepackage{url}            
\usepackage{booktabs}       
\usepackage{amsfonts}       
\usepackage{nicefrac}       
\usepackage{microtype}      
\usepackage{xcolor}         

\definecolor{dgreen}{RGB}{65, 171, 93}

\hypersetup{
    colorlinks=true,
    linkcolor=blue,
    filecolor=magenta,      
    urlcolor=cyan,
    citecolor=dgreen,
    }     

\usepackage{bookmark}
\usepackage{mathtools}
\usepackage{mathalfa}
\usepackage{mathrsfs}
\usepackage{bm,bbm}
\usepackage{ulem}
\usepackage{textcomp}
\usepackage{graphicx}
\usepackage{wrapfig}
\usepackage{gensymb}
\usepackage{floatrow}
\usepackage{enumerate}
\usepackage{algorithmic}
\usepackage{algorithm}
\usepackage{soul}
\usepackage{tikz}
\usepackage{tkz-euclide}
\usepackage{appendix}

\usepackage{chngcntr}
\counterwithin{figure}{section}

\usepackage{amssymb,amsmath,amsthm}
\usepackage{thmtools, thm-restate}

\numberwithin{equation}{section}

\input m.macros
\input m.symbols

\usepackage{natbib}
\setcitestyle{round}
\renewcommand{\bibname}{References}
\renewcommand{\bibsection}{\subsubsection*{\bibname}}

\usepackage[capitalize]{cleveref}

\title{Confronting Reward Model Overoptimization \\ with Constrained RLHF}

\author{Ted Moskovitz\thanks{Corresponding author: \texttt{ted@gatsby.ucl.ac.uk}\newline 
$\quad^{\dagger}$Additional affiliations: Strategy Robot, Inc., Strategic Machine, Inc., Optimized Markets, Inc.}  \\
Gatsby Unit, UCL\\
\centering
\And
\hspace{-6.2cm}Aaditya K. Singh \\
\hspace{-6.2cm}Gatsby Unit, UCL \\
\centering
\AND
DJ Strouse \\
Google DeepMind \\
\centering
\And
Tuomas Sandholm \\
Carnegie Mellon University\footnote{Strategy Robot, Strategic Machine, Optimized Markets}\\
\centering
\And
Ruslan Salakhutdinov \\
Carnegie Mellon University \\
\centering
\AND
Anca D. Dragan \\
University of California, Berkeley \\
\centering
\And 
Stephen McAleer \\
Carnegie Mellon University \\
\centering
}

\begin{document}

\maketitle

\begin{abstract}
Large language models are typically aligned with human preferences by optimizing \textit{reward models} (RMs) fitted to human feedback. However, human preferences are multi-faceted, and it is increasingly common to derive reward from a composition of simpler reward models which each capture a different aspect of language quality. This itself presents a challenge, as it is difficult to appropriately weight these component RMs when combining them. Compounding this difficulty, because any RM is only a proxy for human evaluation, this process is vulnerable to \textit{overoptimization}, wherein past a certain point, accumulating higher reward is associated with worse human ratings. In this paper, we perform, to our knowledge, the first study on overoptimization in composite RMs, showing that correlation between component RMs has a significant effect on the locations of these points. We then introduce an approach to solve this issue using constrained reinforcement learning as a means of preventing the agent from exceeding each RM's threshold of usefulness. Our method addresses the problem of weighting component RMs by learning dynamic weights, naturally expressed by Lagrange multipliers. As a result, each RM stays within the range at which it is an effective proxy, improving evaluation performance. Finally, we introduce an adaptive method using gradient-free optimization to identify and optimize towards these points during a single run. 
\end{abstract}


\section{Introduction}
In the last several years, \textit{Large Language Models} (LLMs) have made impressive advances in natural language processing. These models, which are typically pretrained on massive amounts of text data from the Internet to predict the next token given the current context, are often known as \textit{foundation models} \citep{bommasani2021opportunities} for their ability to be adapted to a variety of downstream applications, such as chatbots \citep{brown2020language,openai2023gpt4,touvron2023llama} or code generation \citep{ahmad2021unified,wang2021codet5,roziere2023code}. This adaptation, or \textit{finetuning}, is often performed via \textit{reinforcement learning from human feedback} \citep[RLHF;][]{knox2008tamer,christiano2017deep,stiennon2020learning}. RLHF treats the pretrained language model as a decision-making agent whose ``actions'' are tokens and whose goal is to maximize a \textit{reward model} (RM) trained to emulate human preferences over output text. As these models become more prevalent in society, there are many concerns regarding their safe deployment \citep{Hendrycks_Hinton_Bengio_Altman_Sutskever_Gates_Grimes_2023,bubeck2023sparks,legg2008machine},
including biases against marginalized or underrepresented groups \citep{bender2021dangers}, proliferation of false information \citep{lin2021truthfulqa}, and leakage of sensitive information \citep{carlini2021extracting}. These concerns are collectively known as the \textit{alignment problem}: how can we ensure that the behavior of these models is aligned with human preferences? 


Current approaches to alignment within RLHF center around the collection of vast amounts of human rating data and the training of larger, more powerful RMs  \citep{ouyang2022training,gao2022scaling}. However, a fundamental issue with any RM is that ultimately, it is only an imperfect proxy for human preferences. \cite{gao2022scaling} drew attention to this fact, showing that maximizing a reward model beyond a certain point can actually begin to decrease ground truth performance (\textit{i.e.}, lead a text-based agent to produce outputs which are judged as qualitatively worse). This phenomenon is known as \textit{reward model overoptimization}. Examples of overoptimization include producing overly wordy responses or hallucinating information in an effort to give the impression of expertise. One simple, yet  expensive, approach to mitigating this issue is to periodically evaluate the model with fresh human rating throughout finetuning and stop early when ratings decline. 

It is also increasingly common to derive reward from \textit{composite RMs}: fixed combinations of several RMs each designed to capture a different aspect of text quality \citep{ramamurthy2022rl4lms, glaese2022improving,yuan2023rrhf,bakker2022fine,wu2023fine}. Such composite RMs are useful because they allow for more fine-grained measurement of agent behavior and each component can be retrained or swapped out without affecting the others. 
Despite these advantages, this approach also presents its own challenges. Determining the weighting among RMs requires hyperparameter optimization to find the combination that produces the best correlation with ground truth evaluation, and the risk of overoptimization means that the best weighting is contingent on a set training duration. Furthermore, when the reward is constructed from several RMs, information about each individual RM is lost, and the agent cannot attribute changes in reward to any single model. In particular, component rewards may even oppose one another, such as an RM which measures safety (and thus may deny certain user requests) versus another rewarding helpfulness \citep{bai2022training}. Worse, early stopping to avoid overoptimization in composite RMs is problematic, as different components will have different values at which they stop being effective proxies for human evaluation.

In this paper, we propose a simple approach to address these challenges: identify the points of overoptimization, which we term \textit{proxy points}, and then use constrained optimization to ensure that each component RM reaches, but does not exceed, its associated proxy point. Rather than use a fixed weighting among components, our method dynamically adapts a weighting to modulate the influence of each RM on the learning process. 
The core idea behind our approach is to use these constraints to prevent the agent from overoptimizing its (composite) RM beyond the proxy points. 

As in existing methods \citep{gao2022scaling}, we rely on some access to ground-truth queries. We propose two ways of using these queries to identify proxy points. 
In the first approach, we train multiple runs and track each reward model value, periodically querying the ground-truth reward model. This approach then finds an optimal joint proxy point by fitting a surface to this data and maximizing it. While effective, this approach requires multiple runs to fit the surface used to find proxy points. In the second approach, we speed up this process by only using one reinforcement learning run. As this run is training, we can periodically query the ground-truth reward model and use this data to run a derivative-free optimization algorithm to find the next candidate proxy points. 
To summarize, we make the following contributions:
\begin{itemize}
    \item 
    We provide 
    analysis of reward model overoptimization in the context of composite reward functions, showing that the correlation between RMs has a significant influence on proxy points.
    \item We propose several constrained RL approaches which incorporate these points into the optimization objectives, preventing overoptimization and improving evaluation performance. 
    \item We show that a derivative-free optimization method can be used to dynamically find these proxy points during a single run, significantly saving computation. 
\end{itemize}

\section{Preliminaries: Reinforcement Learning from Human Feedback}
\paragraph{RL Problem Formulation}
In \textit {reinforcement learning} \citep[RL;][]{sutton2018reinforcement}, an agent seeks to take actions in its environment in order to maximize reward. Mathematically, this problem is typically formalized as a \textit{Markov decision process} \citep[MDP;][]{puterman2014markov}, defined as a tuple $\mathcal M \triangleq (\St, \A, P, r, \gamma, \rho)$, where $\St$ is the state space, $\A$ is the action space, $P: \St \times \A \to \mathcal P(\St)$ is the transition kernel (where $\mathcal P(X)$ denotes the set of distributions over $X$), $r: \St \times \A \times \St \to \reals$ is the reward function, $\gamma \in [0, 1)$ is the discount factor, and $\rho \in \mathcal P(\St)$ is the initial state distribution. In practice, the agent's experience is typically broken into discrete segments, or ``episodes'' of maximum length $T$. At the beginning of each episode, the environment resets and an initial state is sampled $s_0 \sim \rho(\cdot)$. At each time step $t=0, 1, \dots, T-1$, the agent selects an action $a_t$ conditioned on its current state $s_t$ using a stationary policy $\pi(a_t|s_t)$, where $\pi: \St \to \mathcal P(\A)$. Each episode can be summarized as a trajectory $\tau = (s_0, a_0, s_1, \dots, s_T)$. The agent's goal is to find a policy with maximum expected \textit{return} $R(\tau)$, where $R(\tau) \triangleq \sum_{t=0}^{T-1} \gamma^t r(s_t, a_t, s_{t+1})$. The expected return under policy $\pi$ is known as the \textit{value} $v^\pi(s) \triangleq \E[R(\tau)|s_0=s]$ or the \textit{action-value} if conditioned on both states and actions $q^\pi(s, a) \triangleq \E[R(\tau)|s_0=s,a_0=a]$. The optimization problem faced by the agent, then, is
$
    \max_{\pi} \ v^\pi,
$
where $v^\pi \triangleq \E_{s_0\sim\rho(\cdot)} v^\pi(s_0)$ is the average value over initial states.

\paragraph{Integrating Human Feedback} 
The origin and nature of the reward is a fundamental question when formalizing a problem using RL. Using human evaluation to delineate good agent behaviors from bad has a history that extends beyond language models. \cite{knox2008tamer} used human ratings of actions to construct a reward model for the game Tetris, while \cite{christiano2017deep} proposed a mechanism for using human feedback to express preferences over trajectories collected in Atari and MuJoCo. In language modeling, each action is viewed as adding a new token to the current context string~\citep{ziegler2019fine, stiennon2020learning, bai2022training, ouyang2022training}, which can be viewed as the state. The LM is then the policy, with action space $\A$ being the vocabulary of possible tokens, and state space $\St$ being the set of all sequences of tokens up to maximum length $T$.
Transitions are deterministic, with each action token simply appended to the current state.
Given a pretrained LM $\pi_0$, RLHF often consists of three stages \citep{casper2023open}: 1) collecting human feedback on model utterances (typically in the form of ranked preference data), 2) training a RM to model score utterances in alignment with human feedback (typically initialized from a separate pretrained LM) and 3) finetuning the LM with RL using the learned RM. While early work in RLHF for LLMs \citep{stiennon2020learning} focused on a single reward model, more recent work has shown performance benefits of using a weighted combination of simpler RMs \citep{wu2023fine}.

\paragraph{Overoptimization} Recently, \cite{gao2022scaling} performed an empirical study of a phenomenon with deep ramifications for alignment: RM overoptimization. Their core finding is that after a certain point, increasing an LLM agent's value with respect to a given RM will actually begin to decrease its quality on the actual preferences it is trying to learn. (\cite{gao2022scaling} use a ``gold standard'' RM to stand in for human ratings for convenience.) The root of this issue is that any RM is only a proxy for the agent's true measuring stick---human evaluation---so as predicted by Goodhart's Law \citep{goodhart1984problems}, an agent trained to maximize it will eventually learn behaviors which the true objective would discourage. Our approach to addressing this issue is based on a simple two-stage process: first, find the points where the available rewards stop being useful proxies, and second, train an agent to only maximize reward up until that point. 

\section{Finding Proxy Points} \label{sect:proxy-pts}

\paragraph{Setting} In order to conduct an in-depth analysis given our available computational resources, we focus on a single setting as a case study: dialogue generation with the DailyDialog \citep{li2017dailydialog} dataset, which consists of transcripts of conversations between humans. As input, the agent receives a snippet of conversation, and from this context, it must predict the next utterance. We describe this setting in detail in \cref{appendix:experimental_details}. As a base LLM, we follow prior work \citep{wu2023fine} and use GPT-2 \citep{radford2019language} here and throughout this paper. For the reward, we use a combination of two component rewards, each meant to capture a different element of desired behavior, to demonstrate our approach most directly. The first, $r^{met}$, is the METEOR score \citep{banerjee2005meteor} between the generated utterance and reference output, which is computed based on a number of features, including word-matching, synonym-matching, and phrasing. The second, $r^{int}$, measures how well the intent of the generated utterance matches that of the reference output. It is computed using a fine-tuned RoBERTa model \citep{liu2019roberta} which classifies text into different ``intent categories'' such as `inform,' `question,' or `direct.' The typical approach \citep{ramamurthy2022rl4lms} is to linearly combine these RMs to form a composite reward:
\begin{align} \label{eq:combined_reward}
    \tilde r_t = \alpha^{met} r^{met}_t + \alpha^{int} r^{int}_t,
\end{align}
where the coefficients $(\alpha^{met}, \alpha^{int})$ are fixed. As is standard in RLHF applied to language models, an additional KL penalty was added to discourage deviation from the initial model $\pi_0$:
\begin{align}
    r_t =  \tilde r_t - \alpha^{\kl}_t \log\frac{\pi(a_t|s_t)}{\pi_0(a_t|s_t)}. 
\end{align}
The coefficient $\alpha_t^{\kl}$ effectively acts as a Lagrange multiplier, increasing if the KL exceeds some threshold and decreasing otherwise. We discuss this in more detail in \cref{sect:appendix_kl}.

\paragraph{Evaluation and Proxy Points} 
\begin{wrapfigure}[23]{r}{0.6\textwidth}
  \centering
  \vspace{-2ex}
  \includegraphics[width=0.99\textwidth]{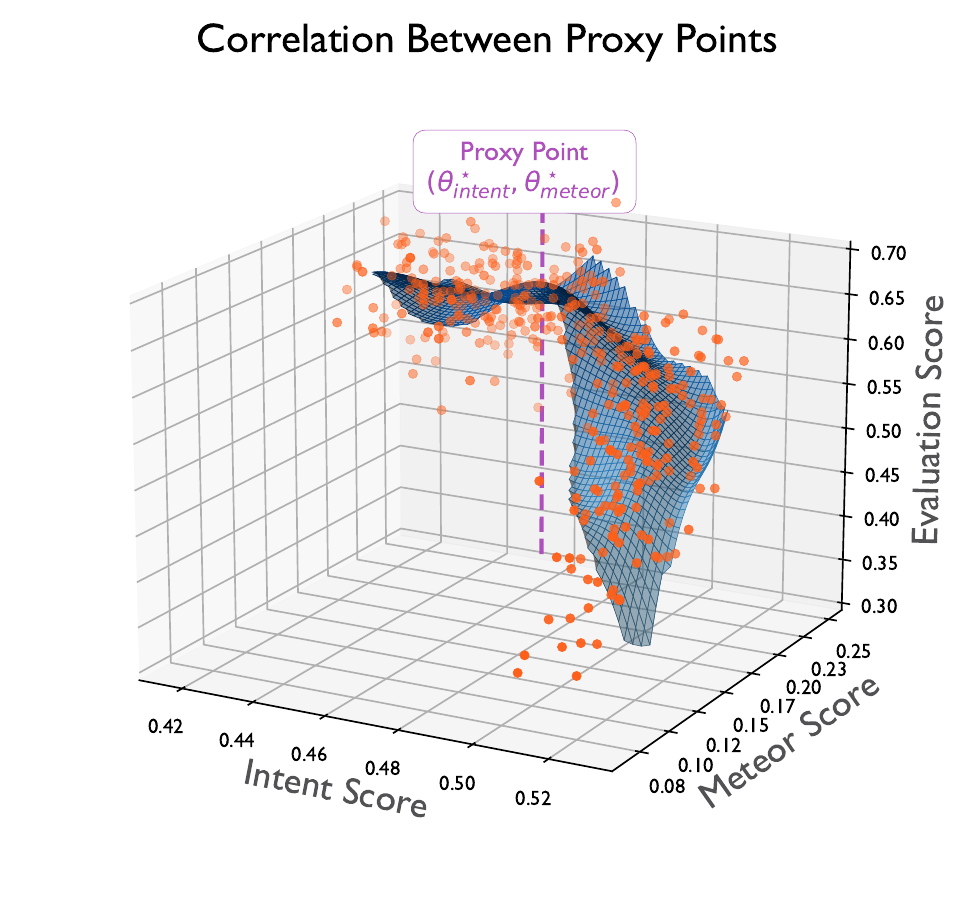}
  \caption{Correlated rewards influence proxy points.}
  \label{fig:correlated}
\end{wrapfigure}
In an ideal world, evaluation performance for all agents across all runs could be measured by collecting a large number of human ratings. However, this is expensive, so we instead selected a number of metrics other than METEOR and intent score which measure the lexical quality and diversity of text outputs and averaged them to serve as our evaluation metric (details in \cref{appendix:experimental_details}). Our choice is in line with prior work that uses held out metrics as the ground truth for convenience of iteration \citep{gao2022scaling}.
We call the value at which further increasing the proxy reward results in decreased ground-truth performance the \textit{proxy point} $\theta^\star$. 
\begin{figure}[t!] 
  \centering
  \includegraphics[width=0.99\textwidth]{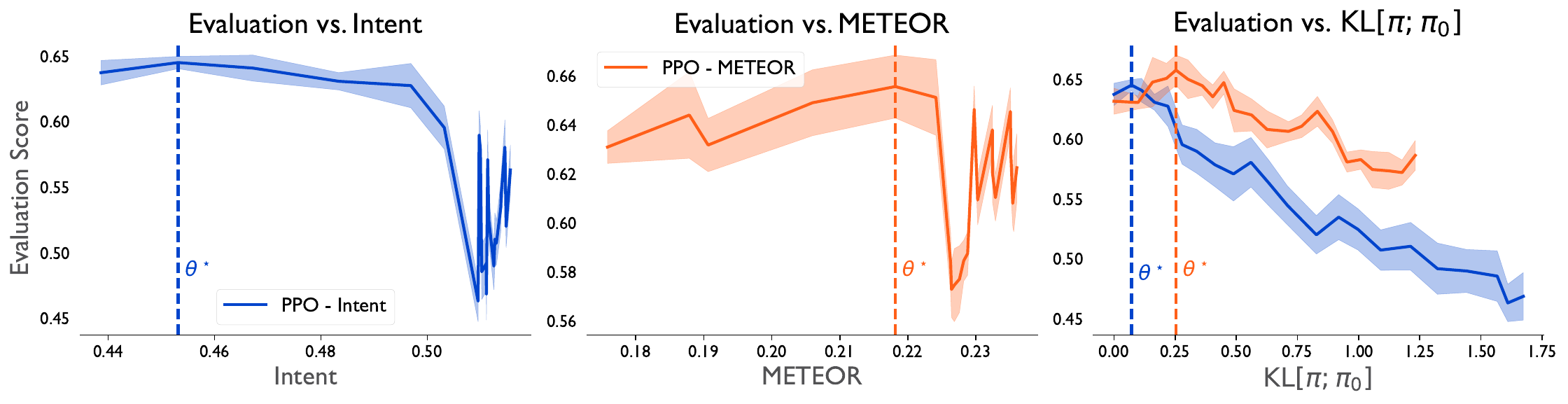}
  \caption{\textbf{Individual RMs are imperfect proxies for evaluation score.} Evaluation score initially increases as individual RMs and the KL divergence grow before falling at proxy points, denoted by dashed lines. Results are averaged over 5 seeds, with shading showing standard error.}
  \label{fig:proxy-pts}
\end{figure}
To identify proxy points, we trained \textsc{PPO} agents \citep{schulman2017ppo} to maximize only one reward or the other (without KL regularization) and plotted the resulting evaluation scores against the METEOR and intent scores in \cref{fig:proxy-pts}. In both cases, the evaluation score initially increases before falling. \cite{gao2022scaling} also observed that, in general, maximization of reward causes the KL divergence between the trained and pretrained policies to increase, and therefore we also expect evaluation score to initially increase before decreasing as the KL grows as well, also shown in \cref{fig:proxy-pts}. One additional phenomenon that makes optimization of composite RMs challenging is that the component RMs may be correlated. We hypothesized that this interaction would influence the proxy points of the component rewards. To test this, we plotted the evaluation scores as a function of the METEOR and intent rewards for each run shown in \cref{fig:proxy-pts} in \cref{fig:correlated} and fit a polynomial surface to the data, using kernel density estimation to only fit the surface over regions with sufficient data (further details in \cref{appendix:experimental_details}). The maximizing point $(\theta_{intent}^\star, \theta_{meteor}^\star)$ indeed differs from the proxy points found by only considering one RM at a time. It is also important to note that the predicted maximizing point is of the fitted surface, rather than any point attained by one of the individual runs. 

\section{Constrained RLHF}
Once one has identified proxy points for the component reward models, the next question is how to train agents to maximize these rewards until they hit these critical values. We propose that a useful approach to doing this is to reformulate the optimization objective using constraints.

\paragraph{Adding Constraints to RL} In constrained reinforcement learning, an agent seeks to maximize its value while adhering to constraints on its behavior. Mathematically, this problem is formalized as a \textit{constrained} MDP \citep[CMDP;][]{altman99cmdps}, which is defined as a tuple $\mathcal M_C \triangleq \left(\St, \A, P, r_0, \gamma, \rho, \{r_{i}\}_{i=1}^N, \{\theta_i\}_{i=1}^N \right)$. Here, $\St$, $\A$, $P$, $r_0$, $\gamma$, and $\rho$ are all as defined for standard MDPs (with $r_0$ the reward function), with $r_i: \St \times \A \to \reals, \ i=1,\dots, N$ being \textit{constraint reward functions} and $\theta_i \in \reals, \ i=1,\dots, N$ associated \textit{constraint thresholds}. Note that the subscripts on $r_{0:N}$ are indices over reward functions, not time steps. For clarity, we will hereafter refer to $r_0$ as the ``task reward'' rather than just the reward. Rather than simply maximize value with respect to $r_0$, the CMDP optimization problem is given by
\begin{align} \label{eq:CMDP_opt}
    \max_\pi \ v^\pi_0 \quad \text{s.t.} \quad v_{i}^\pi \geq \theta_i, \ i=1, \dots, N.
\end{align}
That is, CMDPs represent behaviors which one would like to constrain in the form of value estimates with respect to reward functions which measure these behaviors. The $\geq$ symbol in \cref{eq:CMDP_opt} can easily be reversed if the constraint(s) encode behaviors which should be limited, and the inequality constraint(s) can be replaced with equality constraint(s). While there are many possible formulations, we default to the canonical form in \cref{eq:CMDP_opt} 
for the purposes of exposition. 

\paragraph{Proposed Method} 
Given our possible objectives, we can now consider how to optimize them. One popular approach to solving constrained problems such as \cref{eq:CMDP_opt} is to use Lagrangian relaxation \citep{everett1963lagrangian,altman99cmdps}:
\begin{align} \label{eq:lagrangian}
    \max_\pi \min_{\vec\mu \geq 0}\ v^\pi_0 + \sum_{i=1}^N \mu_i(v_i^\pi - \theta_i) \triangleq \mathcal L(\pi, \vec \mu),
\end{align}
where the weights on the value of each RM $\vec \mu = [\mu_1, \dots, \mu_N]\tr \in \reals_{\geq 0}^N$ are the Lagrange multipliers associated with each constraint. In the case that we use equality constraints rather than inequality constraints, we use the variable $\vec\xi$ rather than $\vec\mu$. Optimization then proceeds by collecting experience using the policy and updating the policy and Langrange multipliers using gradient descent-ascent. We stress that the Lagrange multipliers are \textit{not} fixed hyperparameters, but rather are learned as part of the optimization process. The negative gradient with respect to $\vec \mu$ is simply the constraint violation:
$
    -\grad_{\mu_i} \mathcal L(\pi, \vec \mu) = \theta_i - v_i^\pi
$.
To see how policy optimization works, we can rewrite the Lagrangian as
\begin{align}
\begin{split}
    \mathcal L(\pi, \vec \mu) &=  v_0^\pi + \sum_{i=1}^N \mu_i v_i^\pi - \sum_{i=1}^N \mu_i \theta_i \\
        &= \E_{\stackrel{s_0 \sim \rho(\cdot)}{a_0\sim\pi(\cdot|s_0)}}\left[ q_0^\pi(s_0, a_0) + \sum_{i=1}^N \mu_i q_i^\pi(s_0, a_0) \right]  - \sum_{i=1}^N \mu_i \theta_i 
        \\
        &= \E_{\stackrel{s_0 \sim \rho(\cdot)}{a_0\sim\pi(\cdot|s_0)}}\left[ q_{\vec\mu}^\pi(s_0,a_0)\right] - \sum_{i=1}^N \mu_i \theta_i,
\end{split}
\end{align}
where we define $q_{\vec\mu}^\pi(s,a) \triangleq q_0^\pi(s, a) + \sum_{i=1}^N \mu_i q_i^\pi(s, a)$ as the \textit{mixed} $q$-values of policy $\pi$ given the current Lagrange multipliers $\vec \mu$. Note that this value is non-stationary, as the same policy will have a different value as the weightings on each constraint value change. Policy optimization then proceeds as normal with respect to the mixed $q$-values. As is frequently done in deep RL to reduce variance, we can replace the mixed $q$-values with mixed \textit{advantages} $A_{\vec\mu}^\pi \triangleq q_{\vec\mu}^\pi(s,a) - v_{\vec\mu}(s)$, with $v_{\vec\mu}(s) = \E_{a\sim\pi} q_{\vec\mu}(s,a)$. We can optimize this objective with any policy gradient approach, in our case PPO. 
Detailed pseudocode is provided in \cref{alg:cPPO}. 

\paragraph{Formal Guarantees} While our focus is primarily empirical, we briefly comment on the theoretical properties of the above approach. Lagrangian relaxation converts the CMDP problem into a min-max game. If the values are decomposed as $v^\pi_i = \langle r_i, d_\pi\rangle$, where $d_\pi(s,a) \triangleq (1 - \gamma)\sum_{t\geq 0} \text{Pr}(s_t=s, a_t=a|\pi)$ is the policy's cumulative, discounted state-action occupancy measure, and optimization is performed over $d_\pi$, then the problem is convex-concave and gradient descent-ascent (under basic assumptions) guarantees convergence of the average iterates to a saddle point, \textit{i.e.}, $\left(K^{-1}\sum_{k=1}^K d_\pi^{(k)}, K^{-1}\sum_{k=1}^K\mu^{(k)} \right) \to (d^\star_\pi, \mu^\star)$ as the number of iterations $K \to \infty$ \citep{freund1997online}. However, in large-scale problems it is difficult to optimize directly over $d_\pi$, and we instead update the policy directly. In this case, the problem is convex in $\vec \mu$ but non-concave in $\pi$. \cite{efroni2020cmdps} show sublinear regret bounds with respect to both policy optimality and constraint satisfaction using an optimistic approach, and \cite{ding2020opdop} show a convergence rate for the averaged iterates for general smooth policy classes of $\mathcal O(1/\sqrt{K})$ for the policy and $\mathcal O(1/K^{1/4})$ for the constraint violation using natural policy gradients. 
There is significant work on primal-dual policy optimization for CMDPs, which we discuss further in \cref{sect:appendix_related}. 

\begin{table}[t!]
\begin{center}
\begin{tabular}{lcc}
\toprule
 \textbf{Method}            &     \textbf{Objective} & \textbf{Intuition}  \\
\midrule
 \textsc{PPO} (no KL)   &     $\max_{\pi} \sum_i \alpha_i v_i^\pi$  &     Max. values \\
 \textsc{PPO}  & $\max_{\pi} \sum_i \alpha_i v_i^\pi \ \text{s.t.} \ v_{\kl}^\pi \geq \theta_{\kl}$ & Max. values \& stay close to pretrained $\pi_0$ \\
 \midrule
 \multicolumn{3}{c}{New Methods} \\
 \midrule
  \textsc{PPO}-SAT  & Find $\pi \in \{\pi \vert v_i^\pi = \theta_i \ \forall i \}$ & Find `feasible' policy whose values hit targets \\
 $\mu$-\textsc{PPO} & $\max_{\pi} v_{\kl}^\pi \ \text{s.t.} \ v_{j} \geq \theta_{j} \ \forall j \neq i$ & Stay close to $\pi_0$ \& ensure RMs high enough \\
 All-\textsc{PPO} & $\max_{\pi} \sum_i \alpha_i v_i^\pi \ \text{s.t.} \ v_i \leq \theta_i\ \forall i$ &  Max. RMs but not too much \\
 $\xi$-\textsc{PPO} & $\max_{\pi} v_{\kl}^\pi \ \text{s.t.} \ v_j = \theta_j \ \forall j\neq i$ &  Stay close to $\pi_0$ \& ensure RMs hit targets  \\
\hline
\end{tabular}
\end{center}
\caption{A summary of the approaches we consider.}
\label{table:methods}
\end{table}

\paragraph{Choosing a Constrained Objective} 
Given this approach, we can now consider possible constraint formulations, all of which should embody the intuition that the agent should maximize each component reward only until its corresponding proxy point. This naturally suggests that the proxy points should be used as thresholds in the constrained objective. However, there are a number of possible formulations to consider when casting RLHF as a CMDP with this goal in mind. Once the proxy point for a given RM is reached, the agent has two options: continue to update the Lagrange multiplier on that RM to ensure that values remain at that point (via equality constraints), or simply stop optimizing/un-weight that RM entirely, \textit{i.e.}, set the multiplier to zero, only re-weighting it if the constraint is violated (via inequality constraints). This latter approach carries that risk that the value with respect to that RM will continue to increase (past the proxy point) as other RMs continue to be optimized, but may be empirically effective if this is not the case and optimization is simplified by having a source of non-stationarity eliminated. In both of these cases, each component RM is assigned a constraint threshold, but the question of how to set the task reward remains. We propose the \textit{KL reward} $r_{\kl}= -\log \frac{\pi(a_t|s_t)}{\pi_0(a_t|s_t)}$ as the main task reward. \cite{gao2022scaling} liken the KL to a resource which the agent spends, such that it should try to maximize its reward while limiting its divergence from the original policy as much as possible. Using the negative KL as the task reward carries the intuition of keeping the policy as similar as possible to the pretrained policy, subject to the constraint that each RM hits the point beyond which it stops aligning with the true objective. Note that the requirement that the agent hits these thresholds is crucial, as it prevents the agent from fully maximizing the negative KL reward (\textit{i.e.}, remaining at the pretrained policy). In addition to these, there is another possible constrained approach wherein the agent simply maximizes the combined reward as in standard PPO, but constrained so that each individual RM does not violate its respective threshold. Finally, one could try to formulate the problem as one purely of constraint satisfaction: find any feasible policy whose values with respect to each of the RMs hit the appropriate proxy points. This could be implemented via a reward function that penalizes deviations from these point, \textit{\textit{e.g.}}, $r_{\text{SAT}} = -\sum_i \alpha_i (r_i - \theta_i)^2$. However, this approach faces the same problem as standard PPO---namely, how to best set the weights $\alpha_i$. These proposed approaches are summarized in \cref{table:methods}.

\paragraph{Practical Improvements}
Here, we describe several practical modifications to the ``ideal'' algorithm which we found to improve empirical performance.
In practice, the noise and non-stationarity that primal-dual optimization in RL must contend with can lead to instability in the updates for the Lagrange multipliers. To handle this in practice, we follow prior work \citep{stooke2020pid,zahavy2022domino,moskovitz2023reload}  and use a sigmoid function to bound the Lagrange multipliers between 0 and 1. This results in mixed advantages which are a convex combination of the task and constraint advantages:
\begin{align}
    A_{\vec\mu}^\pi(s,a) = \left(N - \sum_{i=1}^N\sigma(\mu_i)\right) A_0^\pi(s,a) + \sum_{i=1}^N\sigma(\mu_i) A_i^\pi(s,a).
\end{align}
This equation has the intuitive interpretation of placing more weight on optimizing constraint reward $r_{i>0}$ when $\mu_{i>0}$ is high (indicating a constraint violation), and more weight on task reward $r_0$ when $\mu_{1:N}$ are low (indicating that constraints are satisfied). When we use equality constraints rather than inequality constraints, we replace the sigmoid with a $\tanh$ function (bounding the Lagrange multipliers between $-1$ and $1$). When updating the Lagrange multipliers, we found that using low or no momentum in the optimizer (we use SGD with a momentum parameter of 0.1) was helpful for performance, as otherwise $\sigma(\mu_i)$ or $\tanh(\xi_i)$ could be overly ``sticky,'' remaining high for too long when constraints became satisfied and vice versa. Another hack which we found to be useful was to replace the value estimates in the constraint violation calculations with the sum of rewards to-go (for the appropriate reward function) for the remainder of a given episode. This is because we found that early in training, value estimates are inaccurate, which can cause the agent to incorrectly believe it is either adhering to or violating the constraint, leading to incorrect weighting of rewards via the Lagrange multiplier and slower overall learning.

\section{Experimental Evaluation} \label{sect:experiments}

We now evaluate these possible approaches in the same setting as described in \cref{sect:proxy-pts}. The primary questions we would like to answer are as follows. (1) Do constrained methods result in better evaluation performance compared to PPO (and PPO-SAT)? (2) Do these approaches successfully enforce the desired constraints? (3) Do the thresholds determined by the proxy points lead to the best performance? Unless otherwise noted, all experiments are run for 5 random seeds, and any shading in plots denotes standard error. Code for all methods is available here: \url{github.com/tedmoskovitz/ConstrainedRL4LMs}.


\begin{figure}[t!] 
  \centering
  \includegraphics[width=0.95\textwidth]{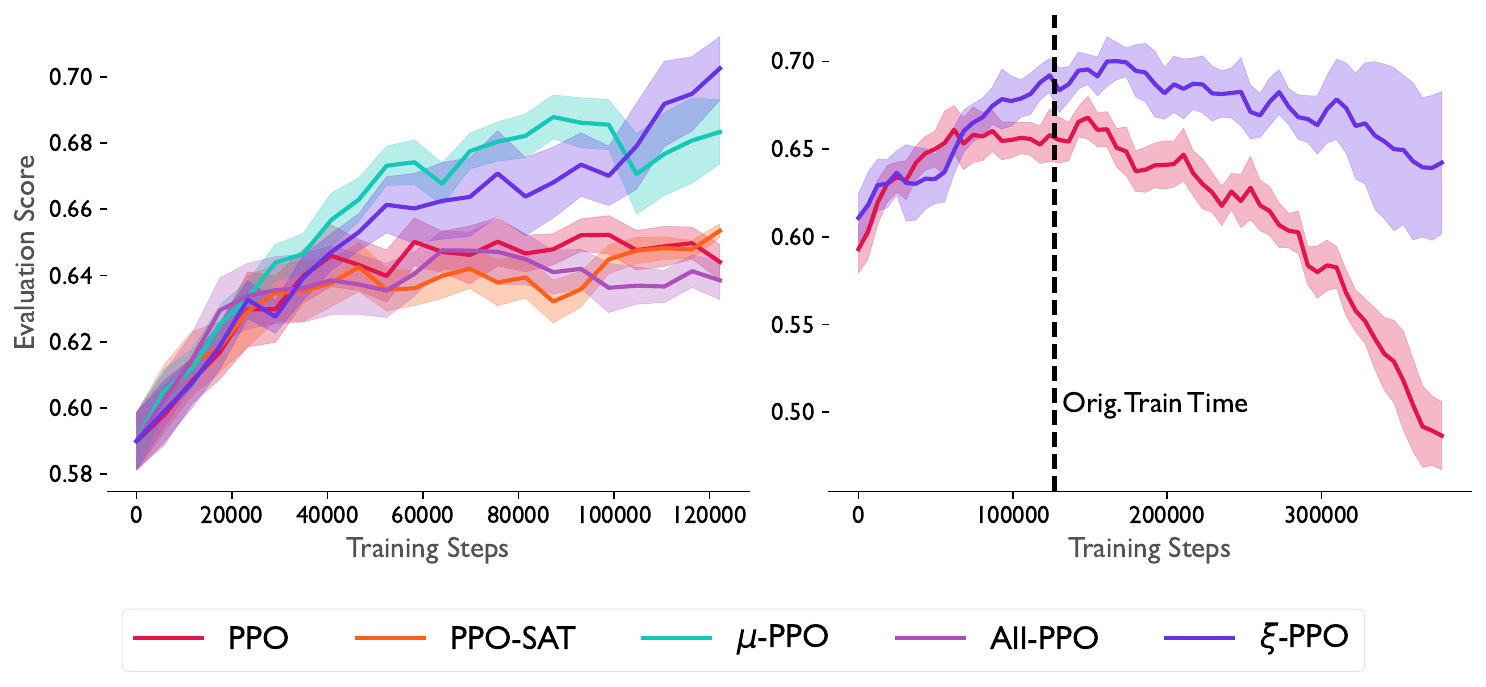}
  \caption{\textbf{Constrained RLHF improves evaluation performance.} (Left) Two constrained methods, $\mu$-PPO and $\xi$-PPO produce the best performance over the course of training. (Right) Balancing RMs using constraints makes performance more robust to longer training time.}
  \label{fig:eval_all}
\end{figure}

\paragraph{Does constrained RLHF improve performance?}
In \cref{fig:eval_all}, we indeed find that two constrained approaches, $\mu$-PPO and $\xi$-PPO achieve better evaluation performance than other methods, with $\xi$-PPO performing slightly better at the end of training. To ensure fairness across methods, to set the fixed RM weightings used to train PPO and PPO-SAT, we selected the best settings found after 10 initial runs of each approach, the same as the total number of runs used to find proxy points used for the constrained methods. We conjecture that the strong performance of $\mu$- and $\xi$-PPO is due to the beneficial effects of jointly optimizing the policy and Lagrange multipliers (RM weightings). For example, even setting the weightings to be the \textit{optimal} Lagrange multipliers and fixing them throughout training is not guaranteed to converge to a saddle point \citep{szepesvari2020cmdps}, a phenomenon observed empirically by \cite{moskovitz2023reload}. Notably, All-PPO did not perform as well as the other constrained methods, which we believe was due to increased instability in the optimization process (Appendix \cref{fig:oscillations}). This is common in constrained problems with ``paradoxical'' objectives \citep{moskovitz2023reload}. Another benefit of continually modulating the weightings among RMs is that the weightings themselves are not hyper-optimized to a particular training duration. We trained both PPO and $\xi$-PPO using their hyperparameter settings optimized over runs with 128,000 steps for 3 times as long over 3 seeds and confirmed that the constrained approach was more stable (\cref{fig:eval_all}). 

\begin{figure}[t!] 
  \centering
  \includegraphics[width=0.8\textwidth]{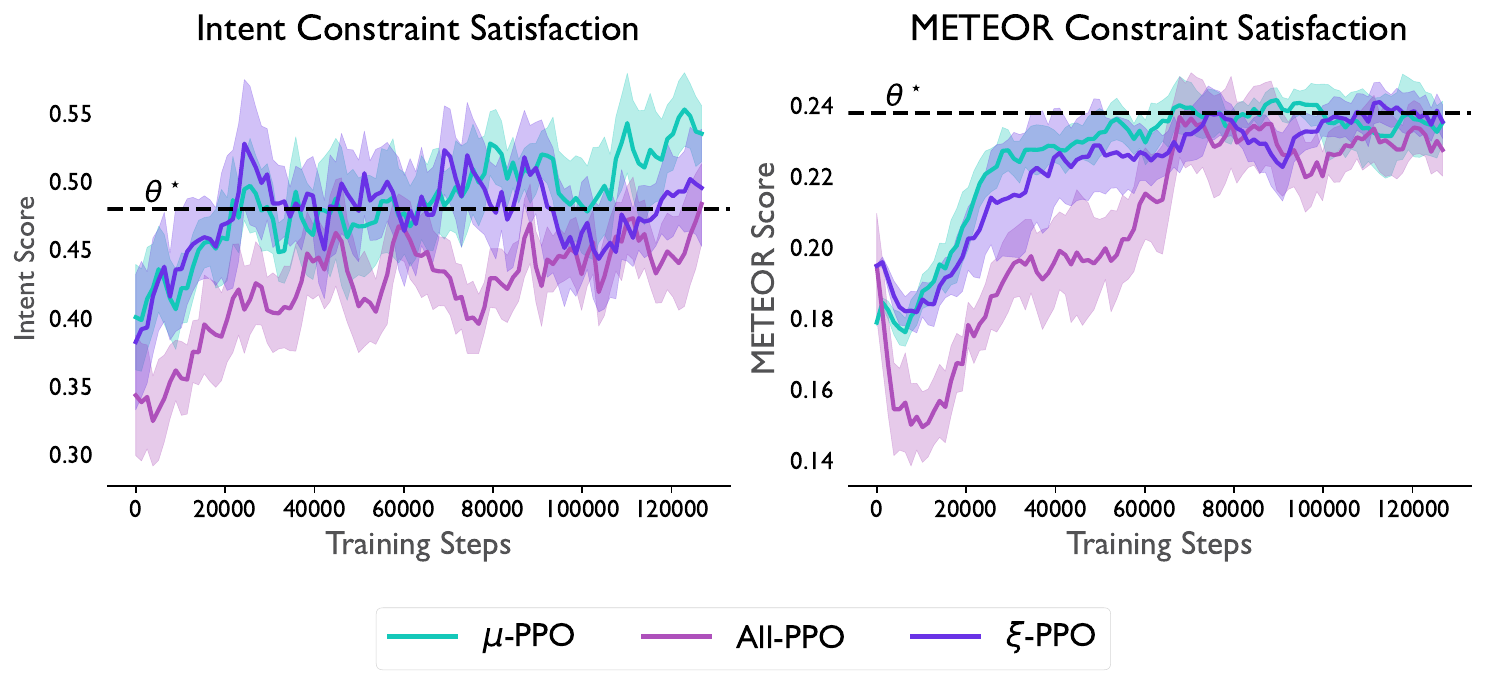}
  \caption{\textbf{Constraints are satisfied.} $\mu$-PPO reaches or exceeds the required intent (left) and METEOR (right) thresholds (dashed lines), All-PPO remains below them, and $\xi$-PPO hits them.}
  \label{fig:constraint_satisfaction}
\end{figure}

\paragraph{Are constraints successfully enforced?}
To verify that the constrained algorithms are working as expected, we plotted the intent and METEOR rewards across training for $\mu$-PPO, All-PPO, and $\xi$-PPO in \cref{fig:constraint_satisfaction}. We can see that, as required by the constraints, $\mu$-PPO (approximately) reaches at least as high as the proxy point thresholds, All-PPO remains below them, and $\xi$-PPO approximately hits them. $\mu$-PPO continues to increase above the intent proxy point, which may contribute to its slightly worse final performance compared to $\xi$-PPO in \cref{fig:eval_all}.

\begin{figure}[t!] 
  \centering
  \includegraphics[width=0.9\textwidth]{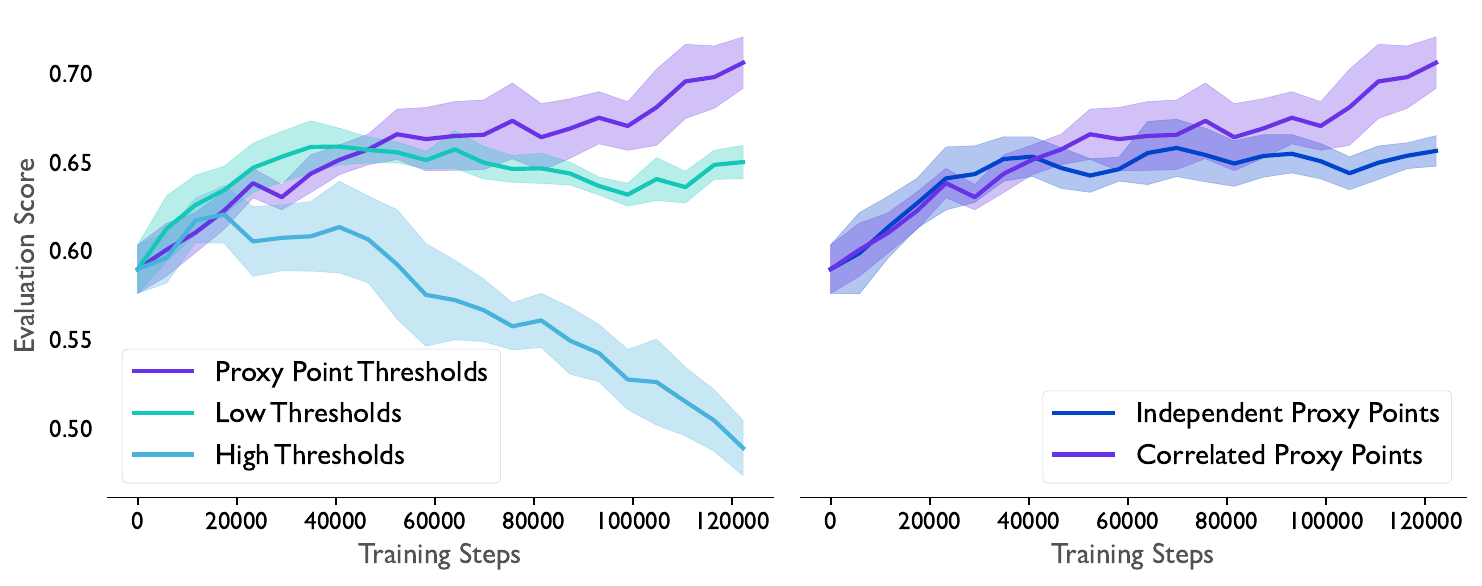}
  \caption{\textbf{Using proxy points as thresholds leads to the best performance.} (Left) Using thresholds that are 10\% lower or higher reduces performance compared to proxy point thresholds. (Right) The proxy points that account for the correlation between RMs are more effective than those estimated independently.}
  \label{fig:proxy-pt_performance}
\end{figure}

\paragraph{Are proxy points the best thresholds?}
We compared the performance of $\xi$-PPO using the proxy points identified in \cref{sect:proxy-pts} against the same method using thresholds that were 10\% lower and 10\% higher. The left panel of \cref{fig:proxy-pt_performance} shows that making thresholds lower causes initial performance to increase more quickly, as once the easier-to-reach thresholds are met, the agent is able to begin tightening the KL with respect to the pretrained policy earlier. However, performance plateaus at a lower level. When thresholds are set too high, the KL reward is ignored and the proxy rewards are optimized beyond the point at which they are useful, leading to worse performance. We also compared the performance of $\xi$-PPO using the correlated proxy points found in \cref{fig:correlated} against the independent proxy points found by only considering one RM at a time (\cref{fig:proxy-pts}). 

\subsection{Improving Threshold Identification}
\begin{figure}[t!] 
  \centering
  \includegraphics[width=0.9\textwidth]{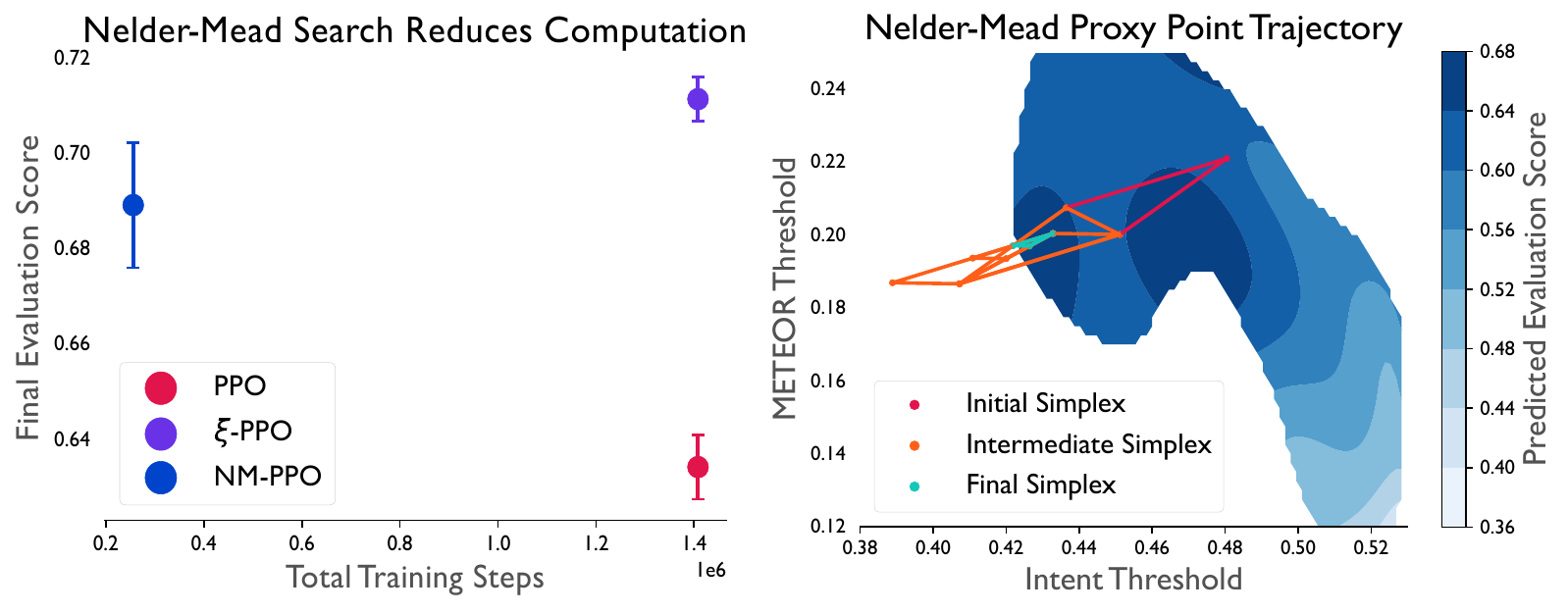}
  \caption{\textbf{Nelder-Mead threshold search saves computation.} (Left) Final evaluation performance versus total number of training steps (including hyperparameters searches). We allowed NM-PPO twice as many training steps for a single run, 256,000. (Right) An example threshold simplex trajectory overlaid on a contour plot of predicted evaluation performance from \cref{fig:correlated}. The search converges to a local maximum.}
  \label{fig:nm}
\end{figure}
One downside of all methods considered so far is the need for multiple runs 
to either select a fixed weighting of RMs or identify proxy points.
It would save significant compute---and reduce environmental impact, particularly for larger models---if it were possible to identify thresholds over the course of a single training run. Assuming we are allowed a limited number of queries to the evaluation metric over the course of training, one approach to accomplishing this would be to use a gradient-free optimizer to update the constraint thresholds to reach better performance. In order to limit the required number of policy updates between threshold updates, we used a local hill-climbing algorithm, Nelder-Mead \citep{nelder1965simplex}, which iteratively updates a simplex of thresholds based on the evaluation performance at each point. Once a new set of thresholds is proposed, we use $\xi$-PPO to converge to those points and then evaluate the model once they're reached. Details are provided in \cref{appendix:nm}. We plotted the final evaluation performance of this variant of our approach, which we term NM-PPO, versus total number of training steps (including runs used for hyperparameter optimization) of PPO and $\xi$-PPO in \cref{fig:nm}. We found that NM-PPO obtains strong performance over the course of a single run, significantly saving in computation. Furthermore, the trajectories of simplexes proposed by Nelder-Mead closely follow the predicted evaluation performance found in \cref{fig:correlated}, converging to local maxima of the surface. In \cref{fig:nm}, the trajectory converges to a local maximum rather than the global maximum, though other runs did indeed find the global optimum as predicted by \cref{fig:correlated} (Appendix \cref{fig:nm_traj2}). 
One caveat with respect to this result is that the feasible region of threshold pairs is relatively small. There is therefore a moderate chance that the initial simplex already contains at least one threshold pair which produces reasonable performance. Further experimentation is required on problems with larger feasible regions and more than two component RMs.

\section{Discussion}

In this work, we studied reward model overoptimization and the influence of correlation on proxy points in composite RMs. Then, we introduced a set of approaches for identifying and using these points as thresholds within a constrained optimization approach to RLHF. One weakness shared by all approaches---unconstrained and constrained alike---is that at least some minimal degree of access to the true objective/evaluation metric is required.  Though in resource-rich settings this could be feasible (\textit{\textit{e.g.}}, by occasionally freezing training and querying human evaluators), ideally, this would be dispensed with entirely. However, doing so is nontrivial. One weakness of gradient descent-ascent applied to primal-dual policy optimization is that it does not guarantee that the final policy and Lagrange multiplier(s) converge to a saddle point, only their averages. It would be an interesting direction for future work to apply an approach which does have such guarantees, such as ReLOAD \citep{moskovitz2023reload}. For optimizing the constraint thresholds during a single run, it would be interesting to explore alternative optimizers to Nelder-Mead, such as Bayesian optimization. Another interesting direction for future work would be to study the usefulness of a CMDP formulation for avoiding degeneration/collapse of model outputs, as while a deterministic optimal policy always exists for standard MDPs, CMDPs may demand optimal policies which are stochastic \citep{szepesvari2020cmdps}. A similar idea was explored using a maximum entropy formulation by \cite{khalifa2020distributional}. In general, further testing of our methods is necessary on more domains and with composite RMs with more components. We believe there are additional interesting avenues to explore in mitigating overoptimization, such as multi-objective RL \citep{abdolmaleki2020dmpo} or with constraints added to supervised learning \citep{rafailov2023direct}. More broadly, we believe constrained optimization offers an important toolbox for approaching the alignment problem.





\paragraph{Acknowledgements}
Ted Moskovitz is funded by the Gatsby Charitable Foundation. Tuomas Sandholm is supported by the Vannevar Bush Faculty Fellowship ONR N00014-23-1-2876, National Science Foundation grants RI-2312342 and RI-1901403, and ARO award W911NF2210266. Stephen McAleer is funded by a CI Fellowship. The authors would like to thank Vivek Veeriah, Tom Zahavy, Misha Laskin, and Dave Abel for helpful discussions.

\clearpage
\bibliographystyle{unsrtnat} 
\bibliography{crlhf_iclr}

\begin{thebibliography}{81}
\providecommand{\natexlab}[1]{#1}
\providecommand{\url}[1]{\texttt{#1}}
\expandafter\ifx\csname urlstyle\endcsname\relax
  \providecommand{\doi}[1]{doi: #1}\else
  \providecommand{\doi}{doi: \begingroup \urlstyle{rm}\Url}\fi

\bibitem[Bommasani et~al.(2021)Bommasani, Hudson, Adeli, Altman, Arora, von
  Arx, Bernstein, Bohg, Bosselut, Brunskill,
  et~al.]{bommasani2021opportunities}
Rishi Bommasani, Drew~A Hudson, Ehsan Adeli, Russ Altman, Simran Arora, Sydney
  von Arx, Michael~S Bernstein, Jeannette Bohg, Antoine Bosselut, Emma
  Brunskill, et~al.
\newblock On the opportunities and risks of foundation models.
\newblock \emph{arXiv preprint arXiv:2108.07258}, 2021.

\bibitem[Brown et~al.(2020)Brown, Mann, Ryder, Subbiah, Kaplan, Dhariwal,
  Neelakantan, Shyam, Sastry, Askell, et~al.]{brown2020language}
Tom Brown, Benjamin Mann, Nick Ryder, Melanie Subbiah, Jared~D Kaplan, Prafulla
  Dhariwal, Arvind Neelakantan, Pranav Shyam, Girish Sastry, Amanda Askell,
  et~al.
\newblock Language models are few-shot learners.
\newblock \emph{Advances in neural information processing systems},
  33:\penalty0 1877--1901, 2020.

\bibitem[OpenAI(2023)]{openai2023gpt4}
OpenAI.
\newblock Gpt-4 technical report, 2023.

\bibitem[Touvron et~al.(2023)Touvron, Martin, Stone, Albert, Almahairi, Babaei,
  Bashlykov, Batra, Bhargava, Bhosale, et~al.]{touvron2023llama}
Hugo Touvron, Louis Martin, Kevin Stone, Peter Albert, Amjad Almahairi, Yasmine
  Babaei, Nikolay Bashlykov, Soumya Batra, Prajjwal Bhargava, Shruti Bhosale,
  et~al.
\newblock Llama 2: Open foundation and fine-tuned chat models.
\newblock \emph{arXiv preprint arXiv:2307.09288}, 2023.

\bibitem[Ahmad et~al.(2021)Ahmad, Chakraborty, Ray, and
  Chang]{ahmad2021unified}
Wasi~Uddin Ahmad, Saikat Chakraborty, Baishakhi Ray, and Kai-Wei Chang.
\newblock Unified pre-training for program understanding and generation.
\newblock \emph{arXiv preprint arXiv:2103.06333}, 2021.

\bibitem[Wang et~al.(2021)Wang, Wang, Joty, and Hoi]{wang2021codet5}
Yue Wang, Weishi Wang, Shafiq Joty, and Steven~C.H. Hoi.
\newblock {C}ode{T}5: Identifier-aware unified pre-trained encoder-decoder
  models for code understanding and generation.
\newblock In \emph{Proceedings of the 2021 Conference on Empirical Methods in
  Natural Language Processing}, pages 8696--8708, Online and Punta Cana,
  Dominican Republic, November 2021. Association for Computational Linguistics.
\newblock \doi{10.18653/v1/2021.emnlp-main.685}.
\newblock URL \url{https://aclanthology.org/2021.emnlp-main.685}.

\bibitem[Rozi{\`e}re et~al.(2023)Rozi{\`e}re, Gehring, Gloeckle, Sootla, Gat,
  Tan, Adi, Liu, Remez, Rapin, et~al.]{roziere2023code}
Baptiste Rozi{\`e}re, Jonas Gehring, Fabian Gloeckle, Sten Sootla, Itai Gat,
  Xiaoqing~Ellen Tan, Yossi Adi, Jingyu Liu, Tal Remez, J{\'e}r{\'e}my Rapin,
  et~al.
\newblock Code llama: Open foundation models for code.
\newblock \emph{arXiv preprint arXiv:2308.12950}, 2023.

\bibitem[Knox and Stone(2008)]{knox2008tamer}
W~Bradley Knox and Peter Stone.
\newblock Tamer: Training an agent manually via evaluative reinforcement.
\newblock In \emph{2008 7th IEEE international conference on development and
  learning}, pages 292--297. IEEE, 2008.

\bibitem[Christiano et~al.(2017)Christiano, Leike, Brown, Martic, Legg, and
  Amodei]{christiano2017deep}
Paul~F Christiano, Jan Leike, Tom Brown, Miljan Martic, Shane Legg, and Dario
  Amodei.
\newblock Deep reinforcement learning from human preferences.
\newblock \emph{Advances in neural information processing systems}, 30, 2017.

\bibitem[Stiennon et~al.(2020)Stiennon, Ouyang, Wu, Ziegler, Lowe, Voss,
  Radford, Amodei, and Christiano]{stiennon2020learning}
Nisan Stiennon, Long Ouyang, Jeffrey Wu, Daniel Ziegler, Ryan Lowe, Chelsea
  Voss, Alec Radford, Dario Amodei, and Paul~F Christiano.
\newblock Learning to summarize with human feedback.
\newblock \emph{Advances in Neural Information Processing Systems},
  33:\penalty0 3008--3021, 2020.

\bibitem[Hendrycks et~al.(2023)Hendrycks, Hinton, Bengio, Altman, Sutskever,
  Gates, and
  Grimes]{Hendrycks_Hinton_Bengio_Altman_Sutskever_Gates_Grimes_2023}
Dan Hendrycks, Geoffrey Hinton, Yoshua Bengio, Sam Altman, Ilya Sutskever, Bill
  Gates, and Grimes.
\newblock Statement on ai risk, May 2023.

\bibitem[Bubeck et~al.(2023)Bubeck, Chandrasekaran, Eldan, Gehrke, Horvitz,
  Kamar, Lee, Lee, Li, Lundberg, et~al.]{bubeck2023sparks}
S{\'e}bastien Bubeck, Varun Chandrasekaran, Ronen Eldan, Johannes Gehrke, Eric
  Horvitz, Ece Kamar, Peter Lee, Yin~Tat Lee, Yuanzhi Li, Scott Lundberg,
  et~al.
\newblock Sparks of artificial general intelligence: Early experiments with
  gpt-4.
\newblock \emph{arXiv preprint arXiv:2303.12712}, 2023.

\bibitem[Legg(2008)]{legg2008machine}
Shane Legg.
\newblock Machine super intelligence.
\newblock \emph{University of Lugano}, 2008.

\bibitem[Bender et~al.(2021)Bender, Gebru, McMillan-Major, and
  Shmitchell]{bender2021dangers}
Emily~M Bender, Timnit Gebru, Angelina McMillan-Major, and Shmargaret
  Shmitchell.
\newblock On the dangers of stochastic parrots: Can language models be too big?
\newblock In \emph{Proceedings of the 2021 ACM conference on fairness,
  accountability, and transparency}, pages 610--623, 2021.

\bibitem[Lin et~al.(2021)Lin, Hilton, and Evans]{lin2021truthfulqa}
Stephanie Lin, Jacob Hilton, and Owain Evans.
\newblock Truthfulqa: Measuring how models mimic human falsehoods.
\newblock \emph{arXiv preprint arXiv:2109.07958}, 2021.

\bibitem[Carlini et~al.(2021)Carlini, Tramer, Wallace, Jagielski, Herbert-Voss,
  Lee, Roberts, Brown, Song, Erlingsson, et~al.]{carlini2021extracting}
Nicholas Carlini, Florian Tramer, Eric Wallace, Matthew Jagielski, Ariel
  Herbert-Voss, Katherine Lee, Adam Roberts, Tom Brown, Dawn Song, Ulfar
  Erlingsson, et~al.
\newblock Extracting training data from large language models.
\newblock In \emph{30th USENIX Security Symposium (USENIX Security 21)}, pages
  2633--2650, 2021.

\bibitem[Ouyang et~al.(2022)Ouyang, Wu, Jiang, Almeida, Wainwright, Mishkin,
  Zhang, Agarwal, Slama, Ray, et~al.]{ouyang2022training}
Long Ouyang, Jeffrey Wu, Xu~Jiang, Diogo Almeida, Carroll Wainwright, Pamela
  Mishkin, Chong Zhang, Sandhini Agarwal, Katarina Slama, Alex Ray, et~al.
\newblock Training language models to follow instructions with human feedback.
\newblock \emph{Advances in Neural Information Processing Systems},
  35:\penalty0 27730--27744, 2022.

\bibitem[Gao et~al.(2022)Gao, Schulman, and Hilton]{gao2022scaling}
Leo Gao, John Schulman, and Jacob Hilton.
\newblock Scaling laws for reward model overoptimization.
\newblock \emph{arXiv preprint arXiv:2210.10760}, 2022.

\bibitem[Ramamurthy et~al.(2022)Ramamurthy, Ammanabrolu, Brantley, Hessel,
  Sifa, Bauckhage, Hajishirzi, and Choi]{ramamurthy2022rl4lms}
Rajkumar Ramamurthy, Prithviraj Ammanabrolu, Kiant{\'e} Brantley, Jack Hessel,
  Rafet Sifa, Christian Bauckhage, Hannaneh Hajishirzi, and Yejin Choi.
\newblock Is reinforcement learning (not) for natural language processing?:
  Benchmarks, baselines, and building blocks for natural language policy
  optimization.
\newblock \emph{arXiv preprint arXiv:2210.01241}, 2022.

\bibitem[Glaese et~al.(2022)Glaese, McAleese, Tr{\k{e}}bacz, Aslanides, Firoiu,
  Ewalds, Rauh, Weidinger, Chadwick, Thacker, et~al.]{glaese2022improving}
Amelia Glaese, Nat McAleese, Maja Tr{\k{e}}bacz, John Aslanides, Vlad Firoiu,
  Timo Ewalds, Maribeth Rauh, Laura Weidinger, Martin Chadwick, Phoebe Thacker,
  et~al.
\newblock Improving alignment of dialogue agents via targeted human judgements.
\newblock \emph{arXiv preprint arXiv:2209.14375}, 2022.

\bibitem[Yuan et~al.(2023)Yuan, Yuan, Tan, Wang, Huang, and
  Huang]{yuan2023rrhf}
Zheng Yuan, Hongyi Yuan, Chuanqi Tan, Wei Wang, Songfang Huang, and Fei Huang.
\newblock Rrhf: Rank responses to align language models with human feedback
  without tears.
\newblock \emph{arXiv preprint arXiv:2304.05302}, 2023.

\bibitem[Bakker et~al.(2022)Bakker, Chadwick, Sheahan, Tessler,
  Campbell-Gillingham, Balaguer, McAleese, Glaese, Aslanides, Botvinick,
  et~al.]{bakker2022fine}
Michiel Bakker, Martin Chadwick, Hannah Sheahan, Michael Tessler, Lucy
  Campbell-Gillingham, Jan Balaguer, Nat McAleese, Amelia Glaese, John
  Aslanides, Matt Botvinick, et~al.
\newblock Fine-tuning language models to find agreement among humans with
  diverse preferences.
\newblock \emph{Advances in Neural Information Processing Systems},
  35:\penalty0 38176--38189, 2022.

\bibitem[Wu et~al.(2023)Wu, Hu, Shi, Dziri, Suhr, Ammanabrolu, Smith,
  Ostendorf, and Hajishirzi]{wu2023fine}
Zeqiu Wu, Yushi Hu, Weijia Shi, Nouha Dziri, Alane Suhr, Prithviraj
  Ammanabrolu, Noah~A Smith, Mari Ostendorf, and Hannaneh Hajishirzi.
\newblock Fine-grained human feedback gives better rewards for language model
  training.
\newblock \emph{arXiv preprint arXiv:2306.01693}, 2023.

\bibitem[Bai et~al.(2022)Bai, Jones, Ndousse, Askell, Chen, DasSarma, Drain,
  Fort, Ganguli, Henighan, Joseph, Kadavath, Kernion, Conerly, El-Showk,
  Elhage, Hatfield-Dodds, Hernandez, Hume, Johnston, Kravec, Lovitt, Nanda,
  Olsson, Amodei, Brown, Clark, McCandlish, Olah, Mann, and
  Kaplan]{bai2022training}
Yuntao Bai, Andy Jones, Kamal Ndousse, Amanda Askell, Anna Chen, Nova DasSarma,
  Dawn Drain, Stanislav Fort, Deep Ganguli, Tom Henighan, Nicholas Joseph,
  Saurav Kadavath, Jackson Kernion, Tom Conerly, Sheer El-Showk, Nelson Elhage,
  Zac Hatfield-Dodds, Danny Hernandez, Tristan Hume, Scott Johnston, Shauna
  Kravec, Liane Lovitt, Neel Nanda, Catherine Olsson, Dario Amodei, Tom Brown,
  Jack Clark, Sam McCandlish, Chris Olah, Ben Mann, and Jared Kaplan.
\newblock Training a helpful and harmless assistant with reinforcement learning
  from human feedback, 2022.

\bibitem[Sutton and Barto(2018)]{sutton2018reinforcement}
Richard~S Sutton and Andrew~G Barto.
\newblock \emph{Reinforcement learning: An introduction}.
\newblock MIT press, 2018.

\bibitem[Puterman(2014)]{puterman2014markov}
Martin~L Puterman.
\newblock \emph{Markov decision processes: discrete stochastic dynamic
  programming}.
\newblock John Wiley \& Sons, 2014.

\bibitem[Ziegler et~al.(2019)Ziegler, Stiennon, Wu, Brown, Radford, Amodei,
  Christiano, and Irving]{ziegler2019fine}
Daniel~M Ziegler, Nisan Stiennon, Jeffrey Wu, Tom~B Brown, Alec Radford, Dario
  Amodei, Paul Christiano, and Geoffrey Irving.
\newblock Fine-tuning language models from human preferences.
\newblock \emph{arXiv preprint arXiv:1909.08593}, 2019.

\bibitem[Casper et~al.(2023)Casper, Davies, Shi, Gilbert, Scheurer, Rando,
  Freedman, Korbak, Lindner, Freire, Wang, Marks, Segerie, Carroll, Peng,
  Christoffersen, Damani, Slocum, Anwar, Siththaranjan, Nadeau, Michaud, Pfau,
  Krasheninnikov, Chen, Langosco, Hase, Bıyık, Dragan, Krueger, Sadigh, and
  Hadfield-Menell]{casper2023open}
Stephen Casper, Xander Davies, Claudia Shi, Thomas~Krendl Gilbert, Jérémy
  Scheurer, Javier Rando, Rachel Freedman, Tomasz Korbak, David Lindner, Pedro
  Freire, Tony Wang, Samuel Marks, Charbel-Raphaël Segerie, Micah Carroll,
  Andi Peng, Phillip Christoffersen, Mehul Damani, Stewart Slocum, Usman Anwar,
  Anand Siththaranjan, Max Nadeau, Eric~J. Michaud, Jacob Pfau, Dmitrii
  Krasheninnikov, Xin Chen, Lauro Langosco, Peter Hase, Erdem Bıyık, Anca
  Dragan, David Krueger, Dorsa Sadigh, and Dylan Hadfield-Menell.
\newblock Open problems and fundamental limitations of reinforcement learning
  from human feedback, 2023.

\bibitem[Goodhart and Goodhart(1984)]{goodhart1984problems}
Charles~AE Goodhart and CAE Goodhart.
\newblock \emph{Problems of monetary management: the UK experience}.
\newblock Springer, 1984.

\bibitem[Li et~al.(2017)Li, Su, Shen, Li, Cao, and Niu]{li2017dailydialog}
Yanran Li, Hui Su, Xiaoyu Shen, Wenjie Li, Ziqiang Cao, and Shuzi Niu.
\newblock Dailydialog: A manually labelled multi-turn dialogue dataset.
\newblock \emph{arXiv preprint arXiv:1710.03957}, 2017.

\bibitem[Radford et~al.(2019)Radford, Wu, Child, Luan, Amodei, Sutskever,
  et~al.]{radford2019language}
Alec Radford, Jeffrey Wu, Rewon Child, David Luan, Dario Amodei, Ilya
  Sutskever, et~al.
\newblock Language models are unsupervised multitask learners.
\newblock \emph{OpenAI blog}, 1\penalty0 (8):\penalty0 9, 2019.

\bibitem[Banerjee and Lavie(2005)]{banerjee2005meteor}
Satanjeev Banerjee and Alon Lavie.
\newblock Meteor: An automatic metric for mt evaluation with improved
  correlation with human judgments.
\newblock In \emph{Proceedings of the acl workshop on intrinsic and extrinsic
  evaluation measures for machine translation and/or summarization}, pages
  65--72, 2005.

\bibitem[Liu et~al.(2019)Liu, Ott, Goyal, Du, Joshi, Chen, Levy, Lewis,
  Zettlemoyer, and Stoyanov]{liu2019roberta}
Yinhan Liu, Myle Ott, Naman Goyal, Jingfei Du, Mandar Joshi, Danqi Chen, Omer
  Levy, Mike Lewis, Luke Zettlemoyer, and Veselin Stoyanov.
\newblock Roberta: A robustly optimized bert pretraining approach.
\newblock \emph{arXiv preprint arXiv:1907.11692}, 2019.

\bibitem[Schulman et~al.(2017)Schulman, Wolski, Dhariwal, Radford, and
  Klimov]{schulman2017ppo}
John Schulman, Filip Wolski, Prafulla Dhariwal, Alec Radford, and Oleg Klimov.
\newblock Proximal policy optimization algorithms.
\newblock \emph{arXiv preprint arXiv:1707.06347}, 2017.

\bibitem[Altman(1999)]{altman99cmdps}
Eitan Altman.
\newblock Constrained markov decision processes, 1999.

\bibitem[Everett(1963)]{everett1963lagrangian}
Hugh Everett.
\newblock Generalized lagrange multiplier method for solving problems of
  optimum allocation of resources.
\newblock \emph{Oper. Res.}, 11\penalty0 (3):\penalty0 399–417, jun 1963.
\newblock URL \url{https://doi.org/10.1287/opre.11.3.399}.

\bibitem[Freund and Schapire(1997)]{freund1997online}
Yoav Freund and Robert~E Schapire.
\newblock A decision-theoretic generalization of on-line learning and an
  application to boosting.
\newblock \emph{Journal of Computer and System Sciences}, 55\penalty0
  (1):\penalty0 119--139, 1997.
\newblock URL
  \url{https://www.sciencedirect.com/science/article/pii/S002200009791504X}.

\bibitem[Efroni et~al.(2020)Efroni, Mannor, and Pirotta]{efroni2020cmdps}
Yonathan Efroni, Shie Mannor, and Matteo Pirotta.
\newblock Exploration-exploitation in constrained mdps, 2020.
\newblock URL \url{https://arxiv.org/abs/2003.02189}.

\bibitem[Ding et~al.(2020)Ding, Wei, Yang, Wang, and Jovanović]{ding2020opdop}
Dongsheng Ding, Xiaohan Wei, Zhuoran Yang, Zhaoran Wang, and Mihailo~R.
  Jovanović.
\newblock Provably efficient safe exploration via primal-dual policy
  optimization, 2020.
\newblock URL \url{https://arxiv.org/abs/2003.00534}.

\bibitem[Stooke et~al.(2020)Stooke, Achiam, and Abbeel]{stooke2020pid}
Adam Stooke, Joshua Achiam, and Pieter Abbeel.
\newblock Responsive safety in reinforcement learning by pid lagrangian
  methods, 2020.
\newblock URL \url{https://arxiv.org/abs/2007.03964}.

\bibitem[Zahavy et~al.(2022)Zahavy, Schroecker, Behbahani, Baumli, Flennerhag,
  Hou, and Singh]{zahavy2022domino}
Tom Zahavy, Yannick Schroecker, Feryal Behbahani, Kate Baumli, Sebastian
  Flennerhag, Shaobo Hou, and Satinder Singh.
\newblock Discovering policies with domino: Diversity optimization maintaining
  near optimality, 2022.
\newblock URL \url{https://arxiv.org/abs/2205.13521}.

\bibitem[Moskovitz et~al.(2023{\natexlab{a}})Moskovitz, O'Donoghue, Veeriah,
  Flennerhag, Singh, and Zahavy]{moskovitz2023reload}
Ted Moskovitz, Brendan O'Donoghue, Vivek Veeriah, Sebastian Flennerhag,
  Satinder Singh, and Tom Zahavy.
\newblock Reload: Reinforcement learning with optimistic ascent-descent for
  last-iterate convergence in constrained mdps.
\newblock \emph{arXiv preprint arXiv:2302.01275}, 2023{\natexlab{a}}.

\bibitem[Szepesv\'ari(2020)]{szepesvari2020cmdps}
Csaba Szepesv\'ari.
\newblock Constrained mdps and the reward hypothesis, Mar 2020.
\newblock URL
  \url{http://readingsml.blogspot.com/2020/03/constrained-mdps-and-reward-hypothesis.html}.

\bibitem[Nelder and Mead(1965)]{nelder1965simplex}
John~A Nelder and Roger Mead.
\newblock A simplex method for function minimization.
\newblock \emph{The computer journal}, 7\penalty0 (4):\penalty0 308--313, 1965.

\bibitem[Khalifa et~al.(2020)Khalifa, Elsahar, and
  Dymetman]{khalifa2020distributional}
Muhammad Khalifa, Hady Elsahar, and Marc Dymetman.
\newblock A distributional approach to controlled text generation.
\newblock \emph{arXiv preprint arXiv:2012.11635}, 2020.

\bibitem[Abdolmaleki et~al.(2020)Abdolmaleki, Huang, Hasenclever, Neunert,
  Song, Zambelli, Martins, Heess, Hadsell, and Riedmiller]{abdolmaleki2020dmpo}
Abbas Abdolmaleki, Sandy Huang, Leonard Hasenclever, Michael Neunert, Francis
  Song, Martina Zambelli, Murilo Martins, Nicolas Heess, Raia Hadsell, and
  Martin Riedmiller.
\newblock A distributional view on multi-objective policy optimization.
\newblock In Hal~Daumé III and Aarti Singh, editors, \emph{Proceedings of the
  37th International Conference on Machine Learning}, volume 119 of
  \emph{Proceedings of Machine Learning Research}, pages 11--22. PMLR, 13--18
  Jul 2020.
\newblock URL \url{https://proceedings.mlr.press/v119/abdolmaleki20a.html}.

\bibitem[Rafailov et~al.(2023)Rafailov, Sharma, Mitchell, Ermon, Manning, and
  Finn]{rafailov2023direct}
Rafael Rafailov, Archit Sharma, Eric Mitchell, Stefano Ermon, Christopher~D
  Manning, and Chelsea Finn.
\newblock Direct preference optimization: Your language model is secretly a
  reward model.
\newblock \emph{arXiv preprint arXiv:2305.18290}, 2023.

\bibitem[Post(2018)]{post2018call}
Matt Post.
\newblock A call for clarity in reporting bleu scores.
\newblock \emph{arXiv preprint arXiv:1804.08771}, 2018.

\bibitem[Lin(2004)]{lin2004rouge}
Chin-Yew Lin.
\newblock Rouge: A package for automatic evaluation of summaries.
\newblock In \emph{Text summarization branches out}, pages 74--81, 2004.

\bibitem[Ganesan(2018)]{ganesan2018rouge}
Kavita Ganesan.
\newblock Rouge 2.0: Updated and improved measures for evaluation of
  summarization tasks.
\newblock \emph{arXiv preprint arXiv:1803.01937}, 2018.

\bibitem[Papineni et~al.(2002)Papineni, Roukos, Ward, and
  Zhu]{papineni2002bleu}
Kishore Papineni, Salim Roukos, Todd Ward, and Wei-Jing Zhu.
\newblock Bleu: a method for automatic evaluation of machine translation.
\newblock In \emph{Proceedings of the 40th annual meeting of the Association
  for Computational Linguistics}, pages 311--318, 2002.

\bibitem[Borkar(2005)]{borkar2005actor}
Vivek~S Borkar.
\newblock An actor-critic algorithm for constrained markov decision processes.
\newblock \emph{Systems \& control letters}, 54\penalty0 (3):\penalty0
  207--213, 2005.

\bibitem[Bhatnagar and Lakshmanan(2012)]{bhatnagar2012online}
Shalabh Bhatnagar and K~Lakshmanan.
\newblock An online actor--critic algorithm with function approximation for
  constrained markov decision processes.
\newblock \emph{Journal of Optimization Theory and Applications}, 153\penalty0
  (3):\penalty0 688--708, 2012.

\bibitem[Achiam et~al.(2017)Achiam, Held, Tamar, and
  Abbeel]{achiam2017constrained}
Joshua Achiam, David Held, Aviv Tamar, and Pieter Abbeel.
\newblock Constrained policy optimization.
\newblock In Doina Precup and Yee~Whye Teh, editors, \emph{Proceedings of the
  34th International Conference on Machine Learning}, volume~70 of
  \emph{Proceedings of Machine Learning Research}, pages 22--31. PMLR, 06--11
  Aug 2017.
\newblock URL \url{https://proceedings.mlr.press/v70/achiam17a.html}.

\bibitem[Chow et~al.(2018)Chow, Nachum, Duenez-Guzman, and
  Ghavamzadeh]{chow2018lyapunov}
Yinlam Chow, Ofir Nachum, Edgar Duenez-Guzman, and Mohammad Ghavamzadeh.
\newblock A lyapunov-based approach to safe reinforcement learning.
\newblock In S.~Bengio, H.~Wallach, H.~Larochelle, K.~Grauman, N.~Cesa-Bianchi,
  and R.~Garnett, editors, \emph{Advances in Neural Information Processing
  Systems}, volume~31. Curran Associates, Inc., 2018.
\newblock URL
  \url{https://proceedings.neurips.cc/paper/2018/file/4fe5149039b52765bde64beb9f674940-Paper.pdf}.

\bibitem[Paternain et~al.(2019)Paternain, Chamon, Calvo-Fullana, and
  Ribeiro]{paternain2019duality}
Santiago Paternain, Luiz Chamon, Miguel Calvo-Fullana, and Alejandro Ribeiro.
\newblock Constrained reinforcement learning has zero duality gap.
\newblock In H.~Wallach, H.~Larochelle, A.~Beygelzimer, F.~d\textquotesingle
  Alch\'{e}-Buc, E.~Fox, and R.~Garnett, editors, \emph{Advances in Neural
  Information Processing Systems}, volume~32. Curran Associates, Inc., 2019.
\newblock URL
  \url{https://proceedings.neurips.cc/paper/2019/file/c1aeb6517a1c7f33514f7ff69047e74e-Paper.pdf}.

\bibitem[Tessler et~al.(2019)Tessler, Mankowitz, and Mannor]{tessler2018reward}
Chen Tessler, Daniel~J. Mankowitz, and Shie Mannor.
\newblock Reward constrained policy optimization.
\newblock In \emph{International Conference on Learning Representations}, 2019.
\newblock URL \url{https://openreview.net/forum?id=SkfrvsA9FX}.

\bibitem[Calian et~al.(2020)Calian, Mankowitz, Zahavy, Xu, Oh, Levine, and
  Mann]{calian2020metal}
Dan~A. Calian, Daniel~J. Mankowitz, Tom Zahavy, Zhongwen Xu, Junhyuk Oh, Nir
  Levine, and Timothy Mann.
\newblock Balancing constraints and rewards with meta-gradient d4pg, 2020.
\newblock URL \url{https://arxiv.org/abs/2010.06324}.

\bibitem[Ding and Lavaei(2023)]{ding2023provably}
Yuhao Ding and Javad Lavaei.
\newblock Provably efficient primal-dual reinforcement learning for cmdps with
  non-stationary objectives and constraints.
\newblock In \emph{Proceedings of the AAAI Conference on Artificial
  Intelligence}, volume~37, pages 7396--7404, 2023.

\bibitem[Padakandla et~al.(2020)Padakandla, KJ, and
  Bhatnagar]{padakandla2020reinforcement}
Sindhu Padakandla, Prabuchandran KJ, and Shalabh Bhatnagar.
\newblock Reinforcement learning algorithm for non-stationary environments.
\newblock \emph{Applied Intelligence}, 50:\penalty0 3590--3606, 2020.

\bibitem[Cheung et~al.(2020)Cheung, Simchi-Levi, and
  Zhu]{cheung2020reinforcement}
Wang~Chi Cheung, David Simchi-Levi, and Ruihao Zhu.
\newblock Reinforcement learning for non-stationary markov decision processes:
  The blessing of (more) optimism.
\newblock In \emph{International Conference on Machine Learning}, pages
  1843--1854. PMLR, 2020.

\bibitem[Lecarpentier and Rachelson(2019)]{lecarpentier2019non}
Erwan Lecarpentier and Emmanuel Rachelson.
\newblock Non-stationary markov decision processes, a worst-case approach using
  model-based reinforcement learning.
\newblock \emph{Advances in neural information processing systems}, 32, 2019.

\bibitem[Khetarpal et~al.(2022)Khetarpal, Riemer, Rish, and
  Precup]{khetarpal2022towards}
Khimya Khetarpal, Matthew Riemer, Irina Rish, and Doina Precup.
\newblock Towards continual reinforcement learning: A review and perspectives.
\newblock \emph{Journal of Artificial Intelligence Research}, 75:\penalty0
  1401--1476, 2022.

\bibitem[Xie et~al.(2020)Xie, Harrison, and Finn]{xie2020deep}
Annie Xie, James Harrison, and Chelsea Finn.
\newblock Deep reinforcement learning amidst lifelong non-stationarity.
\newblock \emph{arXiv preprint arXiv:2006.10701}, 2020.

\bibitem[Xie et~al.(2021)Xie, Harrison, and Finn]{xie2021deep}
Annie Xie, James Harrison, and Chelsea Finn.
\newblock Deep reinforcement learning amidst continual structured
  non-stationarity.
\newblock In \emph{International Conference on Machine Learning}, pages
  11393--11403. PMLR, 2021.

\bibitem[O'Donoghue(2023)]{o2023efficient}
Brendan O'Donoghue.
\newblock Efficient exploration via epistemic-risk-seeking policy optimization.
\newblock \emph{arXiv preprint arXiv:2302.09339}, 2023.

\bibitem[Tarbouriech et~al.(2023)Tarbouriech, Lattimore, and
  O'Donoghue]{tarbouriech2023probabilistic}
Jean Tarbouriech, Tor Lattimore, and Brendan O'Donoghue.
\newblock Probabilistic inference in reinforcement learning done right.
\newblock In \emph{Sixteenth European Workshop on Reinforcement Learning},
  2023.

\bibitem[Moskovitz et~al.(2021{\natexlab{a}})Moskovitz, Wilson, and
  Sahani]{moskovitz2021first}
Ted Moskovitz, Spencer~R Wilson, and Maneesh Sahani.
\newblock A first-occupancy representation for reinforcement learning.
\newblock In \emph{International Conference on Learning Representations},
  2021{\natexlab{a}}.

\bibitem[Moskovitz et~al.(2023{\natexlab{b}})Moskovitz, Hromadka, Touati,
  Borsa, and Sahani]{moskovitz2023state}
Ted Moskovitz, Samo Hromadka, Ahmed Touati, Diana Borsa, and Maneesh Sahani.
\newblock A state representation for diminishing rewards.
\newblock \emph{arXiv preprint arXiv:2309.03710}, 2023{\natexlab{b}}.

\bibitem[Berner et~al.(2019)Berner, Brockman, Chan, Cheung, Debiak, Dennison,
  Farhi, Fischer, Hashme, Hesse, J{\'{o}}zefowicz, Gray, Olsson, Pachocki,
  Petrov, de~Oliveira~Pinto, Raiman, Salimans, Schlatter, Schneider, Sidor,
  Sutskever, Tang, Wolski, and Zhang]{dota}
Christopher Berner, Greg Brockman, Brooke Chan, Vicki Cheung, Przemyslaw
  Debiak, Christy Dennison, David Farhi, Quirin Fischer, Shariq Hashme, Chris
  Hesse, Rafal J{\'{o}}zefowicz, Scott Gray, Catherine Olsson, Jakub Pachocki,
  Michael Petrov, Henrique~Pond{\'{e}} de~Oliveira~Pinto, Jonathan Raiman, Tim
  Salimans, Jeremy Schlatter, Jonas Schneider, Szymon Sidor, Ilya Sutskever,
  Jie Tang, Filip Wolski, and Susan Zhang.
\newblock Dota 2 with large scale deep reinforcement learning.
\newblock \emph{CoRR}, abs/1912.06680, 2019.

\bibitem[Espeholt et~al.(2018)Espeholt, Soyer, Munos, Simonyan, Mnih, Ward,
  Doron, Firoiu, Harley, Dunning, Legg, and Kavukcuoglu]{impala}
Lasse Espeholt, Hubert Soyer, Remi Munos, Karen Simonyan, Vlad Mnih, Tom Ward,
  Yotam Doron, Vlad Firoiu, Tim Harley, Iain Dunning, Shane Legg, and Koray
  Kavukcuoglu.
\newblock {IMPALA}: Scalable distributed deep-{RL} with importance weighted
  actor-learner architectures.
\newblock In \emph{The International Conference on Machine Learning (ICML)}.
  JMLR, 2018.

\bibitem[Kakade and Langford(2002)]{Kakade_Langford:2002_pfd}
Sham Kakade and John Langford.
\newblock Approximately optimal approximate reinforcement learning.
\newblock In \emph{IN PROC. 19TH INTERNATIONAL CONFERENCE ON MACHINE LEARNING},
  pages 267--274, 2002.

\bibitem[Moskovitz et~al.(2021{\natexlab{b}})Moskovitz, Arbel, Huszar, and
  Gretton]{moskovitz2021efficient}
Ted Moskovitz, Michael Arbel, Ferenc Huszar, and Arthur Gretton.
\newblock Efficient wasserstein natural gradients for reinforcement learning.
\newblock In \emph{International Conference on Learning Representations},
  2021{\natexlab{b}}.
\newblock URL \url{https://openreview.net/forum?id=OHgnfSrn2jv}.

\bibitem[Pacchiano et~al.(2020)Pacchiano, Parker-Holder, Tang, Choromanski,
  Choromanska, and Jordan]{pacchiano2020bgrl}
Aldo Pacchiano, Jack Parker-Holder, Yunhao Tang, Krzysztof Choromanski, Anna
  Choromanska, and Michael Jordan.
\newblock Learning to score behaviors for guided policy optimization.
\newblock In Hal~Daumé III and Aarti Singh, editors, \emph{Proceedings of the
  37th International Conference on Machine Learning}, volume 119 of
  \emph{Proceedings of Machine Learning Research}, pages 7445--7454. PMLR,
  13--18 Jul 2020.
\newblock URL \url{https://proceedings.mlr.press/v119/pacchiano20a.html}.

\bibitem[Schulman et~al.(2015)Schulman, Levine, Abbeel, Jordan, and
  Moritz]{schulman2015trpo}
John Schulman, Sergey Levine, Pieter Abbeel, Michael~I. Jordan, and Philipp
  Moritz.
\newblock Trust region policy optimization.
\newblock In Francis~R. Bach and David~M. Blei, editors, \emph{ICML}, volume~37
  of \emph{JMLR Workshop and Conference Proceedings}, pages 1889--1897.
  JMLR.org, 2015.
\newblock URL \url{http://proceedings.mlr.press/v37/schulman15.html}.

\bibitem[Levine(2018)]{Levine:2018_rlai}
Sergey Levine.
\newblock Reinforcement learning and control as probabilistic inference:
  Tutorial and review, 2018.

\bibitem[Haarnoja et~al.(2018)Haarnoja, Zhou, Abbeel, and
  Levine]{Haarnoja:2018}
Tuomas Haarnoja, Aurick Zhou, Pieter Abbeel, and Sergey Levine.
\newblock Soft actor-critic: Off-policy maximum entropy deep reinforcement
  learning with a stochastic actor, 2018.

\bibitem[Abdolmaleki et~al.(2018)Abdolmaleki, Springenberg, Tassa, Munos,
  Heess, and Riedmiller]{abdolmaleki2018mpo}
Abbas Abdolmaleki, Jost~Tobias Springenberg, Yuval Tassa, Remi Munos, Nicolas
  Heess, and Martin Riedmiller.
\newblock Maximum a posteriori policy optimisation, 2018.
\newblock URL \url{https://arxiv.org/abs/1806.06920}.

\bibitem[Galashov et~al.(2019)Galashov, Jayakumar, Hasenclever, Tirumala,
  Schwarz, Desjardins, Czarnecki, Teh, Pascanu, and Heess]{Galashov:2019}
Alexandre Galashov, Siddhant~M. Jayakumar, Leonard Hasenclever, Dhruva
  Tirumala, Jonathan Schwarz, Guillaume Desjardins, Wojciech~M. Czarnecki,
  Yee~Whye Teh, Razvan Pascanu, and Nicolas Heess.
\newblock Information asymmetry in kl-regularized {RL}.
\newblock \emph{CoRR}, abs/1905.01240, 2019.
\newblock URL \url{http://arxiv.org/abs/1905.01240}.

\bibitem[Teh et~al.(2017)Teh, Bapst, Czarnecki, Quan, Kirkpatrick, Hadsell,
  Heess, and Pascanu]{Teh:2017_distral}
Yee~Whye Teh, Victor Bapst, Wojciech~Marian Czarnecki, John Quan, James
  Kirkpatrick, Raia Hadsell, Nicolas Heess, and Razvan Pascanu.
\newblock Distral: Robust multitask reinforcement learning.
\newblock In \emph{Proceedings of the 31st International Conference on Neural
  Information Processing Systems}, NIPS'17, page 4499–4509, 2017.

\bibitem[Moskovitz et~al.(2022)Moskovitz, Arbel, Parker-Holder, and
  Pacchiano]{moskovitz2022towards}
Ted Moskovitz, Michael Arbel, Jack Parker-Holder, and Aldo Pacchiano.
\newblock Towards an understanding of default policies in multitask policy
  optimization.
\newblock In \emph{International Conference on Artificial Intelligence and
  Statistics}, pages 10661--10686. PMLR, 2022.

\end{thebibliography}

\clearpage

\begin{appendix}

\section{Experimental Details} \label{appendix:experimental_details}

\subsection{Setting}
We use the same general experimental setting as \cite{ramamurthy2022rl4lms}. The context window was of length 5, and separating the conversations in this way resulted in 35k training, 3k validation, and 3k test utterances. As in \cite{ramamurthy2022rl4lms}, we use top-$k$, $k=20$ sampling for decoding. The inputs to the model are concatenated snippets of human conversation in which changes of speaker are denoted by a special end-of-utterence (<EOU>) token. The intent classifier reward was derived from a finetuned RoBERTa model \citep{liu2019roberta} which awards a score of 1 if the classified intent of the model's utterance matches that of the reference/ground truth utterance and 0 otherwise. 

\subsection{The Evaluation Metric}
As we note in the main text, our objective in constructing an evaluation metric was to find one for which Goodhart's Law holds with respect to both the METEOR and intent reward functions, not to directly model human preferences. We therefore chose three metrics measuring lexical quality and three metrics measuring text diversity from among the metrics available in the RL4LMs codebase published by \cite{ramamurthy2022rl4lms}. Specifically, the lexical metrics we used were SACREBLEU $x_s$ \citep{post2018call}, ROUGE2 $x_r$ \citep{lin2004rouge,ganesan2018rouge}, and BLEU $x_b$ \citep{papineni2002bleu}, and the diversity metrics we used were \texttt{unique-3} $x_u$, \texttt{vocab\_size-3-nopunct} $x_v$, and \texttt{max\_pred\_length-nopunct} $x_m$. For each metric, we individually normalized the score between 0 and 1 (based on the range of observed values across all runs of PPO - METEOR and PPO - Intent), then averaged the resulting lexical scores and resulting diversity scores, before averaging the two average category scores. More precisely: 
\begin{align}
    \texttt{eval\_score} = \frac{1}{2}\left(\frac{x_s + x_r + x_b}{3} + \frac{x_u + x_v + x_m}{3}\right).
\end{align}

\subsection{Fitting the Evaluation Score Surface}
The overall procedure is described in Phase 1 of \cref{alg:2stage}, where $\mathcal F$ is the function class for the evaluation score estimator. In our case, $\mathcal F$ was the space of polynomials of degree 10. To avoid predicting high evaluation scores over regions of the METEOR $\times$ intent space with little or no data points, we employed kernel density estimation with a Gaussian kernel to create a mask which hid parts of the fitted surface over low-density data regions (with a threshold density of 50/square unit). This approach is purely heuristic and could likely be greatly improved on in future work. 

\subsection{Nelder-Mead PPO Details} \label{appendix:nm}
We provide detailed pseudocode of our approach in \cref{alg:nmPPO}. In practice, we found several implementation details to be important for ensuring good performance. First, the initial simplex was crucial. Rather than initialize thresholds randomly across the entire range of possible METEOR and intent values, we initialize them based on random perturbations of the evaluation of the initial/pretrained policy (\textit{i.e.}, what the METEOR and intent scores are at the beginning of finetuning). This was very helpful, as otherwise Nelder-Mead would propose threshold pairs that were effectively not feasible for the policy to achieve, \textit{e.g.}, a very high METEOR threshold with a very low intent threshold. Second, we capped the number of iterations allowed for one evaluation/threshold setting at 1/8 of the total allowed training steps. Without this, the agent would often waste most of its run trying to hit challenging/infeasible thresholds. If the thresholds couldn't be reached in that time, the evaluation score was computed wherever the agent was at that time. Third, the agent cached the eval scores of previously-reached threshold pairs---if Nelder-Mead proposed a threshold pair that had been reached before (or is within a elementwise tolerance of $\pm5\%$ of a previously-reached pair) then it just returns the evaluation score it measured previously rather than updating the policy to return to it. The Nelder-Mead hyperparameters we use are $\alpha=1, \gamma=2, \rho=0.5, \sigma=0.5$---these settings are untuned, and could likely be adjusted to improve performance. 

\subsection{Computational Resources}
All experiments were performed on a single NVIDIA A100 GPU, with each run taking between 8 and 10 hours with the exception of runs for Nelder-Mead PPO, which took approximately 20 hours.

\begin{algorithm}[!t]
	\caption{Constrained PPO for Dialogue Generation}\label{alg:cPPO}
	\begin{algorithmic}[1] 
	    \STATE \textbf{Require}: Dataset $\mathcal D = \{(\vec x^m, \vec y^m)\}_{m=1}^M$, initial policy parameters $\psi^{(1)}$, initial parameters for value functions $\phi_0^{(1)},\dots,\phi_N^{(1)}$, constraint thresholds $\theta_1^{(1)},\dots,\theta_N^{(1)}$, initial Lagrange multipliers $\vec \mu^{(1)}$
            \FOR{step $k=1,\dots,K$}
                \STATE // Sample experience
                \STATE Uniformly sample $M' < M$ contexts $\vec x^{m'} \sim \mathcal U(\mathcal D)$
                \STATE Generate predicted `trajectory' responses \begin{align*}
                    \hat{\vec y}^{m'} = (a_1, \dots, a_T) \sim p_\pi(\vec y) = \prod_{t=1}^{T} \pi(a_t | s_t)
                \end{align*}
                where $s_1 = \vec x^{m'}$
                \STATE Compute generalized advantage estimates:
                    \begin{align*}
                        (\hat A_i)^{(k)}_t = (\delta_i)_t + \gamma \mu (\delta_i)_{t+1} + \cdots + (\gamma\mu)^{T-t+1}(\delta_i)_{T-1}, \quad i = 0, \dots, N,
                    \end{align*}
                    where $(\delta_i)_t \triangleq r_i(s_t, a_t, s_{t+1}, \vec y^{m'}) + \gamma \bar v_i(s_{t+1}) - v_i(s_t)$.
                \STATE Store advantages and trajectories in buffer $\mathcal B$
                \STATE // Update
                \FOR{epoch $\ell=1, \dots, L$}
                \FOR{trajectory batch $\{( (\hat A_{0:N})_b, (\delta_0)_b, \dots, (\delta_N)_b, \hat{\vec y}_b)\}_{b=1}^B \sim \mathcal U(\mathcal B)$ in $\mathcal B$}
                \STATE \textcolor{purple}{Compute mixed advantage estimates:}
                    \begin{align*}
                        \textcolor{purple}{(\hat A_{\vec\mu})_{bt}^{(k)} = \left(N - \sum_{i=1}^N \sigma\left(\mu_i^{(k)}\right) \right)(\hat A_0)_{bt}^{(k)} + \sum_{i=1}^N \sigma\left(\mu_i^{(k)}\right) (\hat A_i)_{bt}^{(k)}}
                    \end{align*}
                \STATE Compute the policy loss:
                    \begin{align*}
                        \mathcal L_{\textsc{PPO}} = -\frac{1}{BT} \sum_{b=1}^B \sum_{t=0}^{T-1} \min\{\rho_{bt}(\psi^{(k)}) (\textcolor{purple}{\hat A_{\vec \mu}})^{(k)}_{bt}, \mathrm{clip}(\rho_{bt}(\psi^{(k)}), 1-\epsilon, 1+\epsilon)(\textcolor{purple}{\hat A_{\vec \mu}})_{bt}^{(k)} \},
                    \end{align*}
                    where $\rho_{bt}(\psi^{(k)}) = \frac{\pi_{\psi^{(k)}}(a_{bt}|s_{bt})}{\pi_{\psi^{(k-1)}}(a_{bt}|s_{bt})}$.
                \STATE Compute the value function losses:
                    \begin{align*}
                        \mathcal L_{v_i} = \frac{1}{BT}\sum_{b=1}^B \sum_{t=0}^{T-1} \frac{1}{2}(\delta_i)_{bt}^2, \quad i=0,\textcolor{purple}{1 \dots, N}
                    \end{align*}
                \STATE Update the policy and value functions via SGD on $\mathcal L_{\textsc{PPO}} + \alpha_v \sum_{i=0}^N\mathcal L_{v_i}$
                \STATE \textcolor{purple}{Update the Lagrange multipliers via SGD on $\mathcal L_\mu$:}
                    \begin{align*}
                        \textcolor{purple}{\mathcal L_{\mu_i} = \frac{1}{BT}\sum_{b=1}^B \sum_{t=0}^{T-1} (v_i(s_{bt}) - \theta_i)\sigma\left(\mu_i^{(k)}\right)}
                    \end{align*}
                \ENDFOR
                \ENDFOR
                \STATE Reset buffer $\mathcal B \gets \varnothing$
            \ENDFOR
	\end{algorithmic}
\end{algorithm}

\begin{algorithm}
    \caption{Two-Phase Approach}\label{alg:2stage}
    \begin{algorithmic}[1]
        \STATE \textbf{Require:} Proxy RMs $r_{1:N} = r_1,\dots, r_N$, Evaluation RM $r^\star$, policy gradient algorithm $\mathsf{Alg}$, constrained algorithm $\mathsf{CAlg}$ (\textit{\textit{e.g.}}, \cref{alg:cPPO})
        \STATE \texttt{Phase 1: Proxy point identification}
        \STATE Evaluation RM dataset $\mathcal D \gets \varnothing$
        \FOR{$i=1,\dots,N$}
            \STATE Fit $\mathsf{Alg}$ on $r_i$, collect $K$ measurements $\{(r_1, \dots, r_N, r^\star)_k\}_{k=1}^K$ across training
            \STATE $\mathcal D \gets \mathcal D \cup \{(r_1, \dots, r_N, r^\star)_k\}_{k=1}^K$
        \ENDFOR
        \STATE Fit evaluation RM predictor $f_r^\star$
            \begin{align*}
                f_r^\star \gets \argmin_{f_r \in \mathcal F} \frac{1}{NK} \sum_{i=1}^{N}\sum_{k=1}^K (r_{ik}^\star - \tilde f_r((r_1)_{ik}, \dots, (r_N)_{ik}))^2
            \end{align*}
        \STATE Proxy point: $\vec \theta^\star \gets \argmax_{r_1,\dots,r_N} f_r^\star(r_1, \dots, r_N)$
        \STATE \texttt{Phase 2: Constrained optimization}
        \STATE $\pi^\star \gets \mathsf{CAlg}(\vec \theta^\star)$
        \STATE \textbf{Return} $\pi^\star$
    \end{algorithmic}
\end{algorithm}

\begin{algorithm}
    \caption{Nelder-Mead Proxy Point Search}\label{alg:nmPPO}
    \begin{algorithmic}[1]
        \STATE \textbf{Require:} Evaluation RM $r_{eval}$, initial simplex thresholds $\{\vec\theta_j \triangleq (\theta_{1:N})_j\}_{j=1}^{N+1}$, reflection coefficient $\alpha$, expansion coefficient $\gamma$, contraction coefficient $\rho$, shrinkage coefficient $\sigma$
        \STATE Fit $\xi$-PPO using initial thresholds, compute $\{v_{eval}^\pi(\vec\theta_j)\}$
        \WHILE{not converged}
            \STATE Sort threshold sets by evaluation score $(\vec\theta_1, \dots, \vec\theta_{N+1})$
            \STATE Compute the centroid of the $N$-best thresholds $\bar{\vec\theta} = \frac{1}{N}\sum_{i=1}^N \vec \theta_i$ 
            \STATE Reflect the worst point
                $
                    \vec\theta_r = \bar{\vec\theta} + \alpha (\bar{\vec\theta} - \vec\theta_{N+1})
                $
            \STATE Fit $\xi$-PPO on the reflected thresholds and compute $v_{eval}^\pi(\vec\theta_r)$
            \IF{$v_{eval}^\pi(\vec\theta_1) \leq v_{eval}^\pi(\vec\theta_r) < v_{eval}^\pi(\vec\theta_N) $}
                \STATE $\vec\theta_{N+1} = \vec\theta_r$
                \STATE \textbf{GOTO} Line 3
            \ENDIF
            \IF{$v_{eval}^\pi(\vec\theta_r) < v_{eval}^\pi(\vec\theta_0)$}
                \STATE Expand: 
                        $
                        \vec\theta_e = \bar{\vec\theta} + \gamma (\vec\theta_r - \bar{\vec\theta})
                        $
                \STATE Fit $\xi$-PPO on the expanded thresholds and compute $v_{eval}^\pi(\vec\theta_e)$
                \IF{$v_{eval}^\pi(\vec\theta_e) < v_{eval}^\pi(\vec\theta_1)$}
                    \STATE $\vec\theta_{N+1} = \vec\theta_e$
                    \STATE \textbf{GOTO} Line 3
                \ELSE
                    \STATE $\vec\theta_{N+1} = \vec\theta_r$
                    \STATE \textbf{GOTO} Line 3
                \ENDIF
            \ENDIF
            \STATE Contract:
                $
                    \vec\theta_c = \bar{\vec\theta} + \rho(\vec\theta_{N+1} - \bar{\vec\theta})
                $
            \STATE Fit $\xi$-PPO on the contracted thresholds and compute $v_{eval}^\pi(\vec\theta_c)$
            \IF{$v_{eval}^\pi(\vec\theta_c) < v_{eval}^\pi(\vec\theta_{N+1})$}
                \STATE $\vec\theta_{N+1} = \vec\theta_c$
                \STATE \textbf{GOTO} Line 3
            \ENDIF
            \STATE Shrink: $\vec\theta_j \gets \vec\theta_1 + \sigma (\vec \theta_j - \vec\theta_1), \quad j=2,\dots,N+1$
        \ENDWHILE
    \end{algorithmic}
\end{algorithm}

\subsection{Algorithm Hyperparameters}
\begin{table}[ht]
\begin{center}
\begin{tabular}{lrrrrr}
\toprule
 \textbf{Hyperparameter}            &     \textbf{PPO} & \textbf{PPO-SAT}   &  $\mathbf{\mu}$\textbf{-PPO} & \textbf{All-PPO}  & $\mathbf{\xi}$\textbf{-PPO} \\
\midrule
 Steps per Update  ($M'$)     &     1,280  &     1,280   &     1,280  &     1,280   &     1,280 \\
 Total Steps ($KM'$)     &    128,000  &    128,000   &    128,000  &    128,000   &    128,000\\
 Batch Size ($B$)              &    64 &    64   &    64 &    64 
 &    64 \\
 Epochs per Update  ($L$)        &       5  &       5   &       5  &       5  &       5\\
 Learning Rate ($\eta$)         & 1e-6    & 1e-6   & 1e-6    & 1e-6  & 1e-6 \\
 Initial KL Coefficient ($\alpha_0$)    &  0.2 &  0.2  &  0.2 &  0.2  &  0.2\\
 Target KL & 0.5 & 0.5   & - & 0.5  & -\\
 Discount Factor ($\gamma$)  & 0.99 & 0.99  & 0.99 & 0.99  & 0.99\\
 GAE $\lambda$ & 0.95 & 0.95  & 0.95 & 0.95  & 0.95\\
 Clip Ratio ($\epsilon$) & 0.2 & 0.2  & 0.2 & 0.2 & 0.2\\ 
 Rollouts Top-$k$  & 20 & 20   & 20 & 20  & 20\\ 
 Value Function Coefficient ($\alpha_v$) & 0.5 & 0.5  & - & - & -\\ 
 METEOR Coefficient ($\alpha^{met}$)  & 0.5 & 0.5  & - & - & -\\
 Intent Coefficient ($\alpha^{int}$)  & 1.0 & 1.0  & - & - & -\\
 METEOR Proxy Point ($\theta_{meteor}^\star$)  & - & -  & 0.23 & 0.23  & 0.23  \\
 Intent Proxy Point ($\theta_{intent}^\star$)  & - & -  & 0.48 & 0.48 & 0.48\\
 METEOR Value Coefficient  & - & - & 0.5 & 0.5 & 0.5 \\
 Intent Value Coefficient  & - & - & 0.5 & 0.5 & 0.5 \\
 KL Value Coefficient      & - & - & 0.2 & - & 0.2 \\ 
 Lagrange Multiplier Function  & - & - & sigmoid & sigmoid & $\tanh$ \\
\hline
\end{tabular}
\end{center}
\caption{Experiment Hyperparameters.}
\label{table:PPO_hyperparams}
\end{table}

\section{The KL Regularization Coefficient} \label{sect:appendix_kl}
As introduced by \cite{ziegler2019fine}, it is common in RLHF with PPO to adapt the KL coefficient $\alpha^{\kl}$ with the following update:
\begin{align*}
    e_t &= \mathrm{clip}\left(\frac{\kl[\pi(\cdot|s_t); \pi_0(\cdot|s_t)] - \theta^{\kl}}{\theta^{\kl}}, -0.2, 0.2\right) \\
    \alpha_{t+1}^{\kl} &= \alpha_{t}^{\kl}(1 + \eta^{\kl} e_t),
\end{align*}
where $\theta^{\kl}$ is a hyperparameter which effectively acts as an upper limit on the KL from the initial policy, and $\eta^{\kl}$ acts like a learning rate. The KL coefficient then follows the path of a Lagrange multiplier with $\theta^{\kl}$ as its constraint threshold, as the constraint violation $\kl[\pi(\cdot|s_t); \pi_0(\cdot|s_t)] - \theta^{\kl}$ is exactly the gradient with respect to such a Lagrange multiplier. 

\section{Additional Related Work}  \label{sect:appendix_related}
In addition to the discussion in the main text, there is a long history of work on CMDPs. \cite{borkar2005actor} first studied actor-critic approaches in this context, and \cite{bhatnagar2012online} were the first to consider constrained policy optimization with function approximation. More broadly, \cite{achiam2017constrained}, \cite{chow2018lyapunov}, \cite{paternain2019duality}, \cite{tessler2018reward}, \cite{calian2020metal}, \cite{efroni2020cmdps}, \cite{stooke2020pid}, \cite{moskovitz2023reload}, and \cite{ding2023provably} all study the problem of integrating constraints into RL. More generally, an important factor in using a Lagrangian approach to solving CMDPs is the introduction of non-stationarity into the reward function. RL with non-stationary rewards is an active area of interest in RL \citep{padakandla2020reinforcement,cheung2020reinforcement,lecarpentier2019non}, particularly in the context of continual RL \citep{khetarpal2022towards} often with some form of temporal structure introduced in the non-stationarity \citep{xie2020deep,xie2021deep}. An interesting case in additional to primal-dual optimization in which non-stationarity is introduced by the agent itself is in the use of epistemic uncertainty for more efficient exploration, manifested in the form of non-stationary exploration bonuses to reward \citep{o2023efficient,tarbouriech2023probabilistic}. Non-stationarity may also be introduced as a means of modeling more naturalistic reward structures for studying animal behavior \citep{moskovitz2021first,moskovitz2023state}. Finally, another area of related work is regularized policy optimization, whereby the standard reward-maximizing policy optimization objective is augmented with a regularization term, typically a divergence measure with respect to some reference policy \citep{dota,impala}. In the single-task setting, the updated policy is typically regularized to stay close to its current setting, which has close connections to natural gradient \citep{Kakade_Langford:2002_pfd,moskovitz2021efficient,pacchiano2020bgrl}, trust region \citep{schulman2015trpo}, and variational inference \citep{Levine:2018_rlai,Haarnoja:2018,abdolmaleki2018mpo} approaches. In the multitask setting, the policy is typically regularized towards some default policy which encodes behavior thought to be useful across a family of tasks, and which may be far from the current policy \citep{Galashov:2019,Teh:2017_distral,moskovitz2022towards}. This setting is quite similar in this sense to KL regularization as used in RLHF.

\section{Additional Results}  \label{sect:appendix_results}

\subsection{Additional Metrics}

\begin{figure}[t!] 
  \centering
  \includegraphics[width=0.99\textwidth]{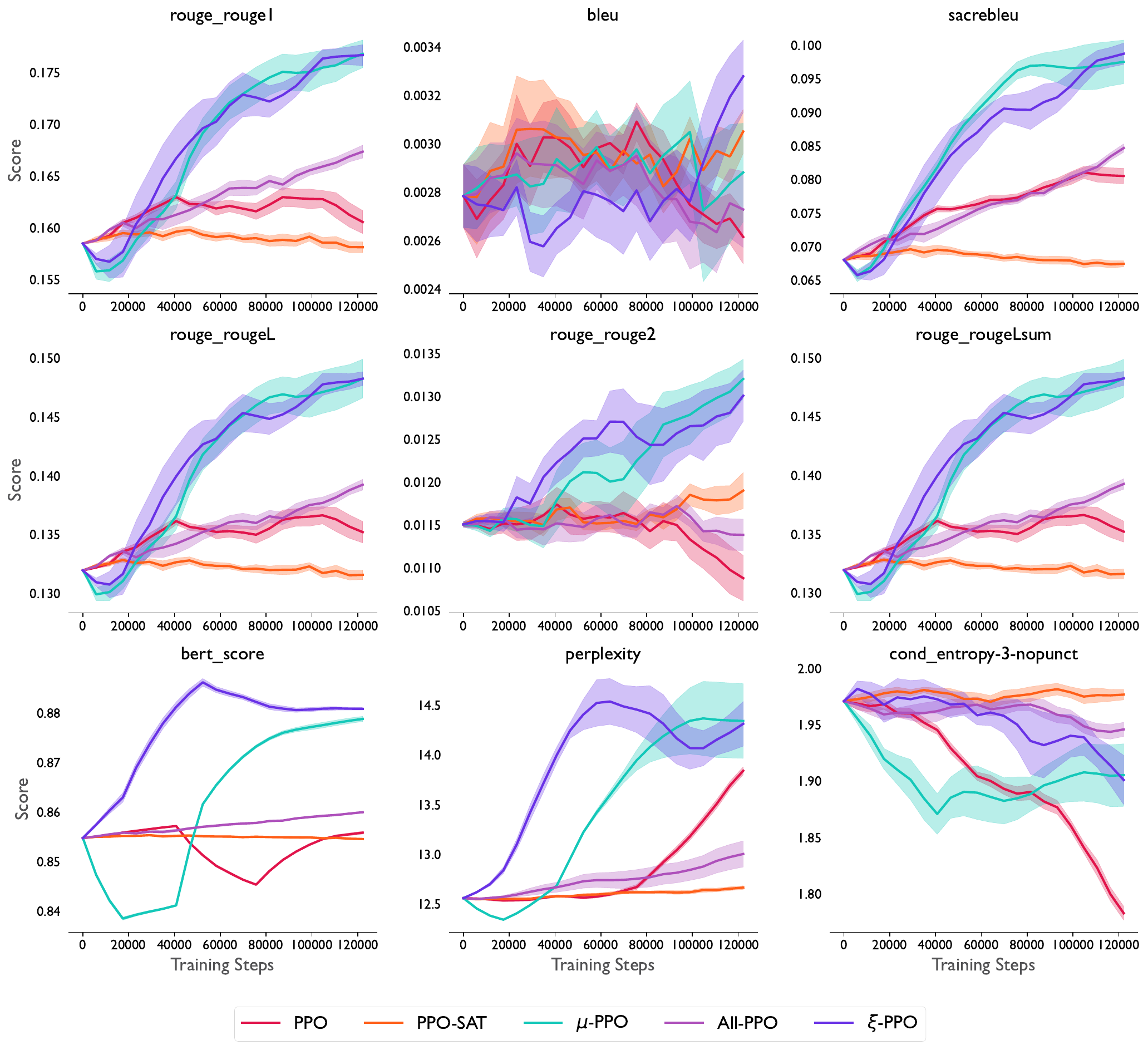}
  \caption{\textbf{Performance of the tested methods across various metrics.}}
  \label{fig:extra_metrics}
\end{figure}

\begin{figure}[t!] 
  \centering
  \includegraphics[width=0.8\textwidth]{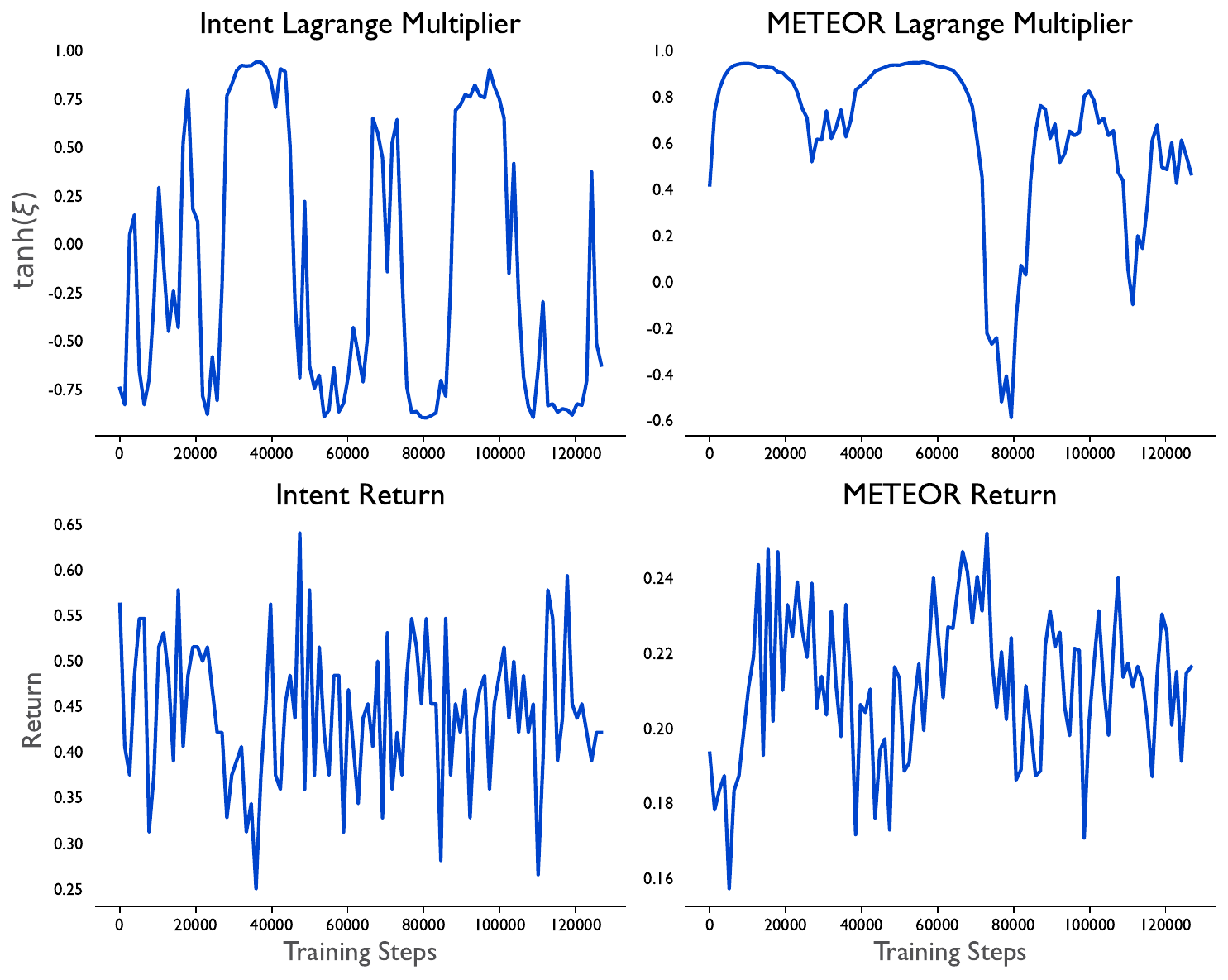}
  \caption{\textbf{Intermediate thresholds can cause oscillations.} Running gradient descent-ascent on a min-max game only guarantees that the average of the iterates converges to the saddle point. In practice, this can mean that the Lagrange multiplier(s) and value(s) can oscillate wildly over the course of training, even if their averages converge. The problem is worse for constraint thresholds which are intermediate---those that are neither high nor low relative to the range of an individual reward function \citep{moskovitz2023reload}, but can be hidden by averaging. Above is an example run of All-PPO, showing this problem can occur. }
  \label{fig:oscillations}
\end{figure}

\begin{figure}[t!] 
  \centering
  \includegraphics[width=0.6\textwidth]{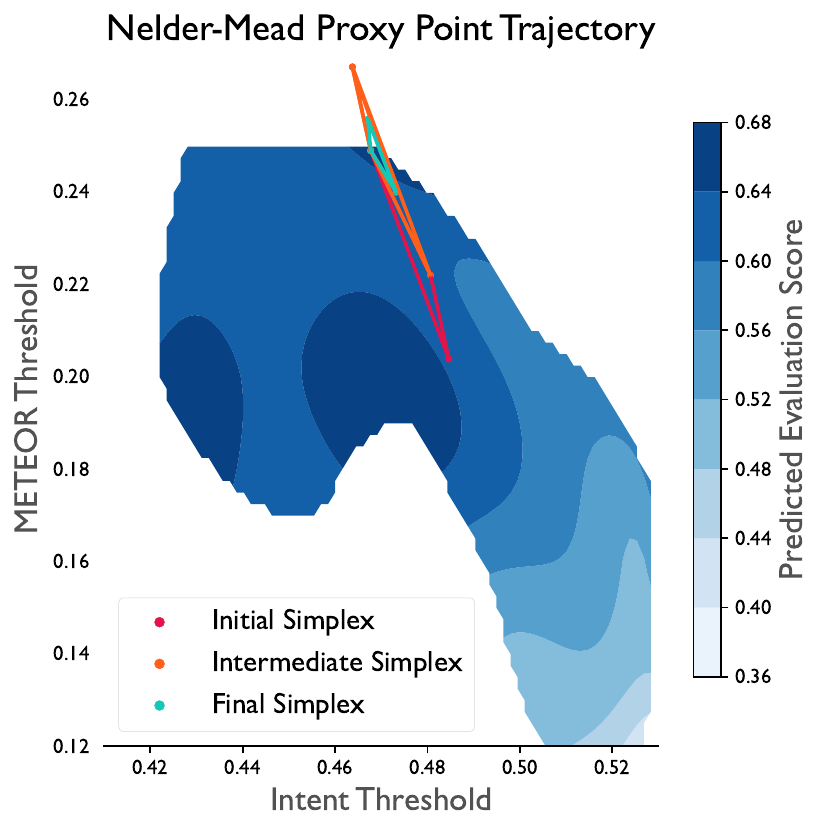}
  \caption{\textbf{Another example Nelder-Mead simplex trajectory.} In this case, Nelder-Mead converged to the global maximum proxy point setting. }
  \label{fig:nm_traj2}
\end{figure}

\begin{figure}[t!] 
  \centering
  \includegraphics[width=0.6\textwidth]{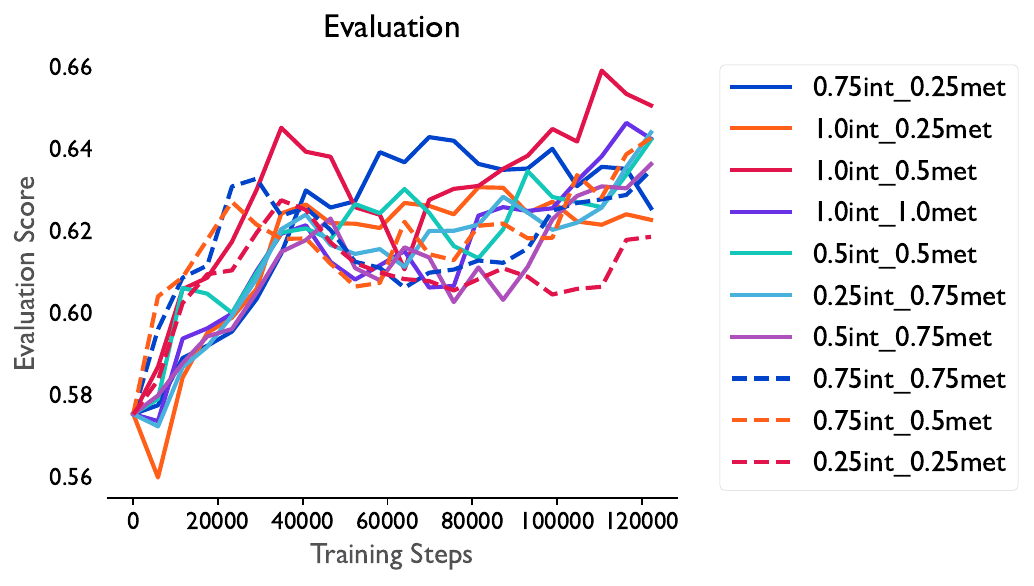}
  \caption{\textbf{Sweep over possible RM weightings for PPO.}}
  \label{fig:ppo_sweep}
\end{figure}

\clearpage
\subsection{Sample Outputs}
For reference, we provide several random prompts and model outputs for the various methods. 

\paragraph{Prompt:} \texttt{ They didn ’ t accept me . <EOU> Are you serious ? <EOU> It ’ s the truth . <EOU> Are you okay ? <EOU> I really wanted to get into NYU .  <EOU>} 

\begin{itemize}
    \item \textbf{Reference:} \texttt{I am really sorry to hear that . <EOU>}
    \item \textbf{PPO:} \texttt{What were you trying to hide.  <EOU> I'm not hiding anything. <EOU>}
    \item \textbf{PPO-SAT:} \texttt{What do you mean? Did something happen to you? <EOU>}
    \item $\mu$\textbf{-PPO:} \texttt{ What do you intend on doing in college? <EOU> I don't know if I'll}
    \item \textbf{All-PPO:} \texttt{What do you mean u don't accept people. <EOU>}
    \item $\xi$\textbf{-PPO:} \texttt{What did you do? <EOU>}
\end{itemize}

\paragraph{Prompt:} \texttt{ I can't . I'm waiting for Paul , but he's late . <EOU> I'll say the movie's starting in the minute . <EOU> I know . I wonder what's keeping him . <EOU> Looks like he won't show up . He's done this before , hasn't he ? <EOU> Yeah , a number of times .  <EOU>} 

\begin{itemize}
    \item \textbf{Reference:} \texttt{What's he trying to pull anyway . <EOU>}
    \item \textbf{PPO:} \texttt{You're right. I think he's been looking for Paul. <EOU> He said.}
    \item \textbf{PPO-SAT:} \texttt{So we're waiting. If you want to stay a bit longer you can go with Paul. <EOU>}
    \item $\mu$\textbf{-PPO:} \texttt{You're looking forward to meeting him tomorrow. <EOU> I'll find out. <EOU>}
    \item \textbf{All-PPO:} \texttt{So we're waiting.. <EOU> And he just didn't show up. <EOU>}
    \item $\xi$\textbf{-PPO:} \texttt{So I'm waiting. <EOU> He's been doing this for a long time. <}
\end{itemize}

\paragraph{Prompt:} \texttt{What's wrong with you ? <EOU> I have a sore throat and headache . <EOU> Do you feel tired ? <EOU> Yes , my whole body feels weak . And I really feel terrible . <EOU> I'm sorry to hear that . I'll have to examine you .  <EOU>} 

\begin{itemize}
    \item \textbf{Reference:} \texttt{Ok , go ahead . <EOU>}
    \item \textbf{PPO:} \texttt{What's the matter? You're always so busy. <EOU>}
    \item \textbf{PPO-SAT:} \texttt{What is your name? <EOU> Moth <EOU> Moth <EOU>}
    \item $\mu$\textbf{-PPO:} \texttt{ What's wrong with your food? <EOU> No matter what.. I'll have to check}
    \item \textbf{All-PPO:} \texttt{What is the most important step of your quest? <EOU> Making sure that your quest succeeds}
    \item $\xi$\textbf{-PPO:} \texttt{ What is your condition? Do you have any fever? <EOU> No}
\end{itemize}

\end{appendix}

\end{document}